\documentclass[journal]{IEEEtran}
\usepackage{ifpdf}

\usepackage{booktabs,amsfonts}
\usepackage{multirow}

\newcommand{\etal}{\textit{et al}.}
\newcommand{\ie}{\textit{i}.\textit{e}., }
\newcommand{\eg}{\textit{e}.\textit{g}., }
\usepackage{color}
\usepackage{amsmath}
\usepackage{xcolor}

\usepackage{cite}

\ifCLASSINFOpdf
   \usepackage[pdftex]{graphicx}
\else
\fi
\usepackage{amsmath}
\usepackage{algorithmic}

\usepackage{array}
\begin{document}
%
\title{Stimuli-Aware Visual Emotion Analysis}
%
%
%

\author{Jingyuan~Yang,~\IEEEmembership{}
	Jie~Li,~\IEEEmembership{}
	Xiumei~Wang,~\IEEEmembership{}
	Yuxuan~Ding~\IEEEmembership{}
	and~Xinbo~Gao,~\IEEEmembership{Senior~Member,~IEEE}
	\thanks{
		Manuscript received January 21, 2020; revised April 6, 2021 and May 11, 2021; accepted
		August 9, 2021. 
		This work was supported in part by the National Natural Science Foundation of China under Grants 62036007, 62050175 and 61772402.
		The associate editor coordinating the review of this manuscript and approving it for publication was Prof. Trac D. Tran. (\textit{Corresponding author: Xinbo Gao.})
		
		J. Yang, J. Li, X. Wang and Y. Ding are with the School of Electronic Engineering, Xidian University, Xi'an 710071, China (e-mail: jingyuanyang@stu.xidian.edu.cn; leejie@mail.xidian.edu.cn; wangxm@xi-dian.edu.cn; yxding@stu.xidian.edu.cn).
		
		X. Gao is with the School of Electronic Engineering, Xidian University, Xi'an 710071, China (e-mail: xbgao@mail.xidian.edu.cn) and with
		the Chongqing Key Laboratory of Image Cognition, Chongqing University of Posts and Telecommunications, Chongqing 400065, China (e-mail:
		gaoxb@cqupt.edu.cn).
		
	}
}

%
%

\markboth{IEEE TRANSACTIONS ON IMAGE PROCESSING}%
{Shell \MakeLowercase{\textit{et al.}}: Bare Demo of IEEEtran.cls for IEEE Journals}
%



\maketitle

\begin{abstract}
	Visual emotion analysis (VEA) has attracted great attention recently, due to the increasing tendency of expressing and understanding emotions through images on social networks. 
	Different from traditional vision tasks, VEA is inherently more challenging since it involves a much higher level of complexity and ambiguity in human cognitive process. 
	Most of the existing methods adopt deep learning techniques to extract general features from the whole image, disregarding the specific features evoked by various emotional stimuli. 
	Inspired by the \textit{Stimuli-Organism-Response (S-O-R)} emotion model in psychological theory, we proposed a stimuli-aware VEA method consisting of three stages, namely stimuli selection (S), feature extraction (O) and emotion prediction (R). 
	First, specific emotional stimuli (\ie color, object, face) are selected from images by employing the off-the-shelf tools. 
	To the best of our knowledge, it is the first time to introduce stimuli selection process into VEA in an end-to-end network. 
	Then, we design three specific networks, \ie Global-Net, Semantic-Net and Expression-Net, to extract distinct emotional features from different stimuli simultaneously. 
	Finally, benefiting from the inherent structure of Mikel's wheel, we design a novel hierarchical cross-entropy loss to distinguish hard false examples from easy ones in an emotion-specific manner.  
	Experiments demonstrate that the proposed method consistently outperforms the state-of-the-art approaches on four public visual emotion datasets.
	Ablation study and visualizations further prove the validity and interpretability of our method.  
\end{abstract}


%
\IEEEpeerreviewmaketitle

\section{Introduction}
\label{sec:introduction}
\begin{figure}
	\centering
	\includegraphics[width=0.85\linewidth]{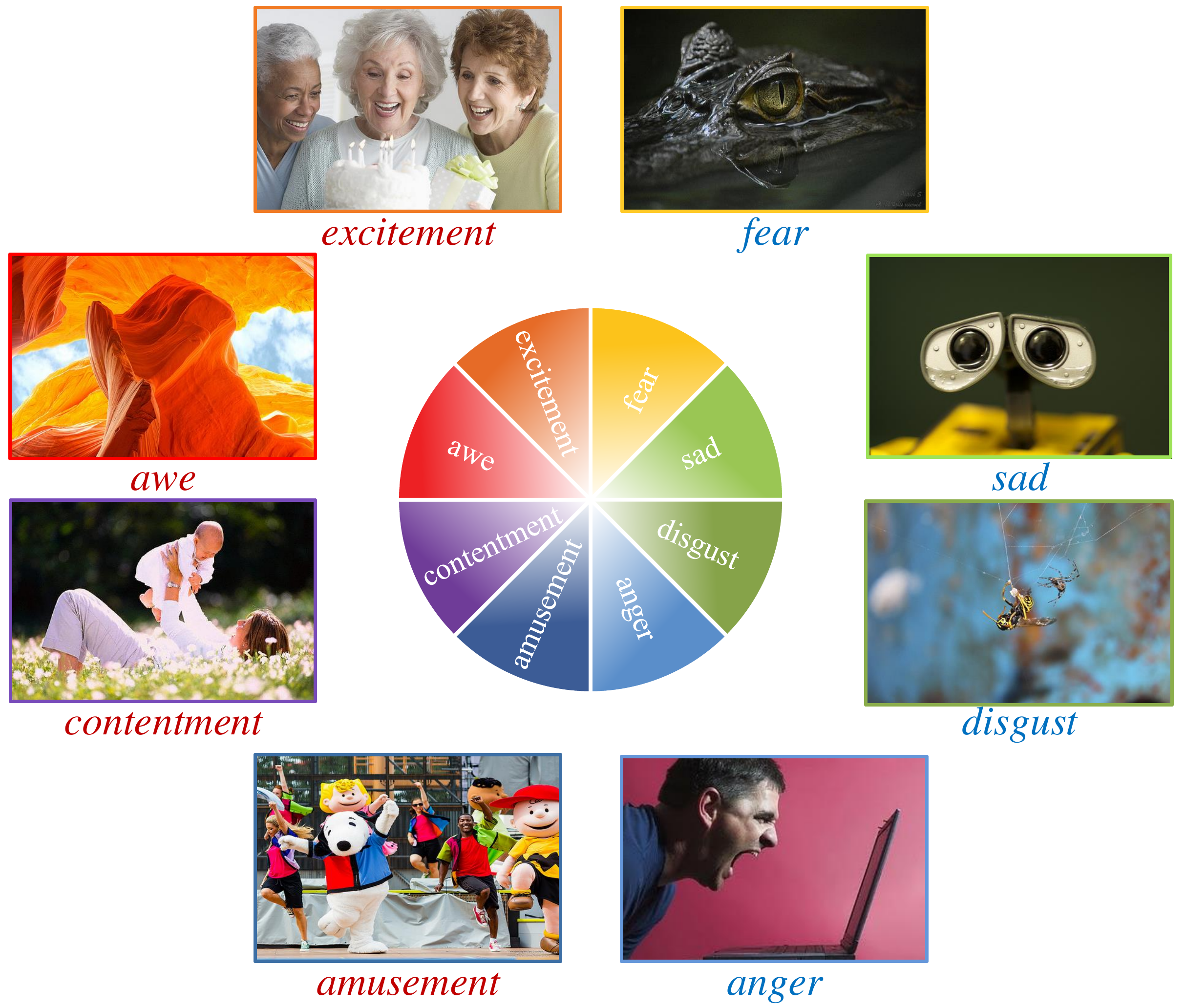}
	\caption{Mikel's wheel from psychological model.
	Affective images from FI dataset with emotion categories (\ie \textit{amusement, anger, awe, contentment, disgust, excitement, fear, and sad}) and polarities (\ie \textit{positive, negative}).
	}
	\label{fig:intro_2}
\end{figure}

In recent years, after the breathtaking success in traditional computer vision tasks such as object detection~\cite{girshick2014rich,girshick2015fast,ren2015faster} and image classification~\cite{krizhevsky2012imagenet, simonyan2014very, he2016deep}, researchers have gradually turned their eyes from low-level vision tasks to high-level ones, including visual reasoning~\cite{li2019visual}, image aesthetic assessment~\cite{kao2017deep}, visual emotion analysis~\cite{yang2018weakly}, \textit{etc.}
Among all the high-level vision tasks, \textit{Visual Emotion Analysis (VEA)} is one of the most challenging tasks for the existing \textit{affective gap}~\cite{zhao2018affective} between low-level pixels and high-level emotions. 
Against all odds, VEA is still promising as understanding human emotions is a crucial step towards strong artificial intelligence~\cite{de2013personal}.
As shown in Fig.~\ref{fig:intro_2}, people may experience different emotions towards different images according to Mikel's wheel~\cite{mikels2005emotional, zhao2016predicting}.
Consequently, VEA aims at finding out how people feel emotionally towards different visual information, which has become an important research topic recently. 
Solving the VEA task may have a tremendous impact on real-world applications such as opinion mining~\cite{binali2009new, yadollahi2017current}, smart advertising~\cite{sanchez2020opinion, chang2001impacts}, and mental disease treatment~\cite{chapman2019borderline, yang2018emhealth}. 

Various methods have been proposed to deal with this challenging yet promising problem, including earlier traditional methods and recent deep learning ones.
In an early stage, inspired by psychological and aesthetic theories, researchers designed assorted hand-crafted features manually, which included color, texture, composition, balance, emphasis, \textit{etc.}~\cite{machajdik2010affective,zhao2014exploring,siersdorfer2010analyzing,zhao2014affective,borth2013large}. 
Most of the early attempts designed specific yet limited features, which failed to cover all important emotional factors and thus resulted in degraded performance on large-scale datasets. 
With the rapid development of \textit{Convolutional Neural Networks (CNNs)}, more and more researchers employed deep learning networks to VEA~\cite{chen2014deep,rao2016learning,yang2018visual,yang2018weakly,yang2018retrieving,yang2017joint,you2015robust}. 
Instead of designing emotion features manually, CNNs were capable of mining emotional representations automatically in an end-to-end manner and consequently achieved better results. 
Nevertheless, most of the existing deep learning methods directly extracted general features from the whole image, which failed to consider the unique evocation process involved in VEA.

In view of the significance of VEA, great attention has been paid to it not only in computer vision but also in psychology. 
Psychologist Frijda~\cite{frijda1986emotions} found that some special substances evoked emotion and named them emotional stimuli.
Psychologists Mehrabian and Russell~\cite{mehrabian1974approach} suggested that people perceived emotions through three steps -- \textit{stimuli}, \textit{organism} and \textit{response}, which was the so-called \textit{S-O-R} model.
In traditional computer vision tasks, networks often consist of two parts, namely feature extraction and prediction, which can be regarded as organism (O) and response (R) in S-O-R model, correspondingly. 
However, S-O-R model suggests that it is specific stimuli rather than the whole image that evoke emotions, which makes stimuli selection (S) indispensable to VEA compared with other vision tasks. 
Therefore, as opposed to extracting features from the whole image directly, we first select emotional stimuli from the image and extract specific features from those stimuli afterwards.

\begin{figure}
	\centering
	\includegraphics[width=0.95\linewidth]{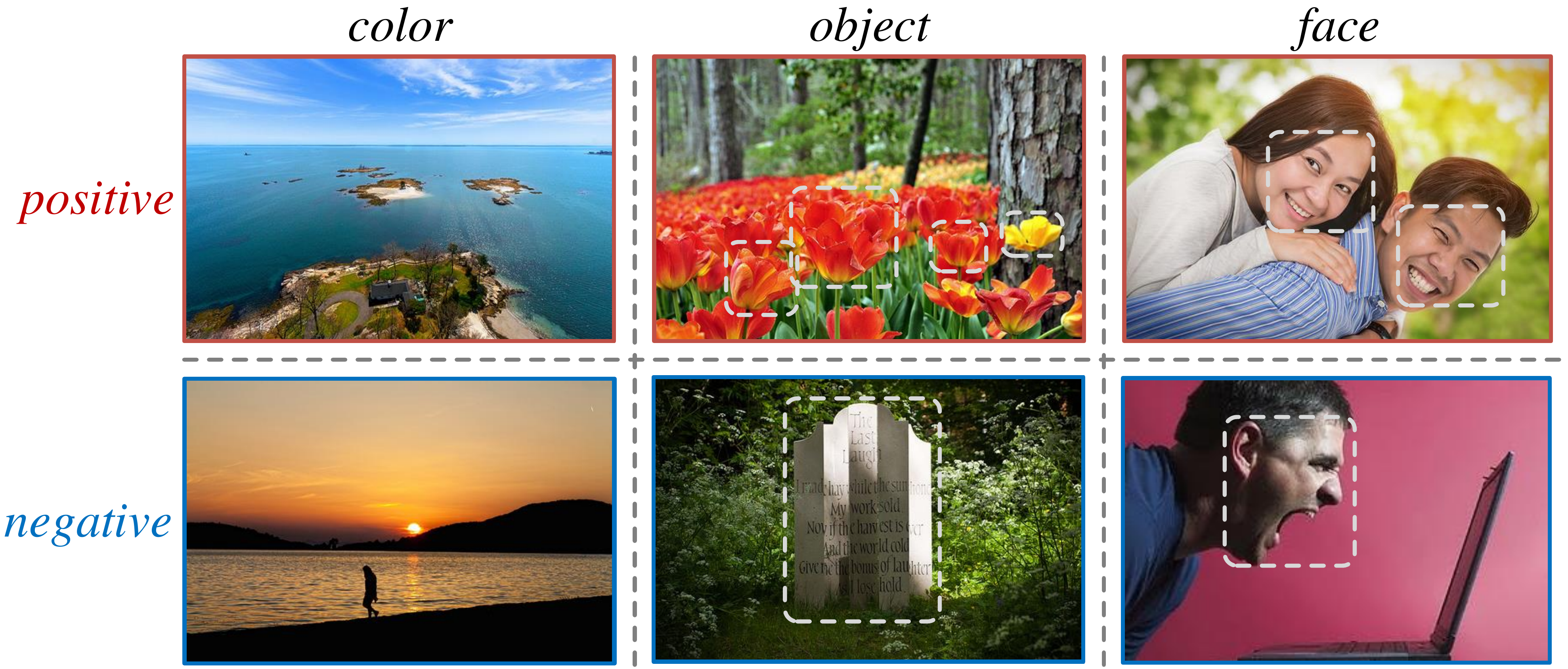}
	\caption{Examples from FI dataset.
		Different emotional stimuli (\ie color, object, face) may evoke different emotions (\ie positive, negative).}
	\label{fig:intro_1}
\end{figure}

As mentioned in a psychological literature~\cite{brosch2010perception}, color, certain objects, facial expressions or any other attributes can be summarized as emotional stimuli.
When it comes to a scenery image without any salient objects, the emotion is evoked by global features, such as color, brightness, texture, \textit{etc.}
When several objects are presented in an image, people tend to deduce the semantic correlations between those objects and subsequently generate an emotion. 
When human face appears in an image, we may focus on the facial expression and evoke a similar emotion as well, which is called empathy~\cite{rogers1975empathic}.
As shown in Fig.~\ref{fig:intro_1}, under different emotional stimuli (\ie color, object, and face), people may experience different emotions.
For example, sunset may bring people a negative feeling while smiling faces often deliver a positive emotion.  
Therefore, in the proposed method, we choose color, object and face as special emotional stimuli both theoretically~\cite{brosch2010perception} and empirically. 

Therefore, motivated by the above facts from psychology, we proposed a novel stimuli-aware VEA network consisting of three stages, namely stimuli selection (S), feature extraction (O) and emotion prediction (R).
In the first stage, special emotional stimuli (\ie color, object and face) are selected by employing the off-the-shelf tools in deep learning.
After that, we design three specific networks to extract distinct features from different stimuli simultaneously, namely Global-Net, Semantic-Net and Expression-Net.
Eventually, we leverage all the above features for emotion prediction.

Categories in most classification tasks are irrelevant to each other, while there exists an inherent hierarchical structure between emotions according Mikel's wheel~\cite{mikels2005emotional, zhao2016predicting}, a widely-used emotion model from psychology.
As shown in Fig.~\ref{fig:intro_2}, emotions on Mikel's wheel are separated into eight categories (\ie \textit{amusement, anger, awe, contentment, disgust, excitement, fear, and sad}) as well as two polarities (\ie \textit{positive and negative}).
To be specific, amusement, awe, contentment, excitement belong to positive emotions while anger, disgust, fear, sad are negative ones, where different colors (\ie red and blue) in Fig.~\ref{fig:intro_2} denote positive and negative emotions respectively.
To exploit such prior knowledge, we propose a novel hierarchical cross-entropy loss for VEA.
In general classification tasks, examples can be divided into true examples and false ones considering whether their emotion categories are correctly classified.
We further distinguish hard false examples from easy false ones by judging whether their emotion polarities are correctly classified.
To be specific, we add an auxiliary polarity loss to the traditional emotion loss, with increased penalties on hard false examples compared with easy ones.
By implementing the proposed hierarchical cross-entropy loss, the whole network can pay more attention to hard false examples and consequently predict emotions in a hierarchical and emotion-specific manner.

Our contributions can be summarized as follows:
\begin{itemize}
	\item We propose a stimuli-aware VEA network consisting of three stages, namely stimuli selection (S), feature extraction (O) and emotion prediction (R), which corresponds to the three stages in S-O-R model from psychology.
	The proposed method consistently outperforms the state-of-the-art methods on four public visual emotion datasets.
	\item We choose color, object and face as special emotional stimuli theoretically and empirically, and design three specific networks to extract features from different stimuli simultaneously.
	To the best of our knowledge, it is the first work that predict emotions from specific emotional stimuli instead of the whole image.
	\item We design a novel hierarchical cross-entropy loss to distinguish hard false examples from easy ones with increased penalties, which benefits from the inherent structure of psychological model and further boosts the classification result in an emotion-specific manner.
\end{itemize}

The rest of the paper is organized as follows. Section~\ref{sec:related_work} overviews the existing methods on visual emotion analysis, object detection and facial expression recognition. In Section~\ref{sec:methodology}, we introduce our proposed stimuli-aware VEA network and the novel hierarchical cross-entropy loss. Extensive experiments and visualization results on four public datasets are given in Section~\ref{sec:experimental_results} and Section~\ref{sec:visualization_results}. Finally, we conclude our work in Section~\ref{sec:conclusion}.

\section{Related work}
\label{sec:related_work}
In this section, we review the previous methods on VEA from earlier traditional methods and recent deep learning ones.
Meanwhile, related work on object detection and facial expression recognition is also included in this section.

\subsection{Visual Emotion Analysis}
\label{sec:visual_emotion_analysis}
Emotion psychologists established two models describing emotions, namely Categorical Emotion States (CES), and Dimensional Emotion Space (DES).
CES model divided emotions into several categories, which mainly included a two-category approach and an eight-category approach. 
DES model employed a continuous Valence-Arousal-Dominance (VAD) space to separate different emotions. 
As for simplicity and intuitiveness, most researchers preferred CES model to DES model. 
Some typical models in CES included Ekman's six emotion categories~\cite{ekman1992argument} and Mikel's eight emotion categories~\cite{mikels2005emotional, zhao2016predicting}, serving as our basic models as well.

\subsubsection{Traditional Methods}
\label{sec:traditional_methods}
Earlier work on VEA mainly focused on the design of traditional hand-crafted features. Machajdik~\etal~\cite{machajdik2010affective} extracted low-level features (\ie color, texture, composition, and content), and combined them to predict emotions of an image.
Siersdorfer~\etal~\cite{siersdorfer2010analyzing} adopted global and local RGB histogram, as well as SIFT-based bag of visual terms as emotion features.
Zhao~\etal~\cite{zhao2014exploring} investigated the relationship between artistic principles and emotions by extracting principle-of-art-based emotion features, which consisted of balance, emphasis, harmony, variety, gradation, and movement. 
Borth~\etal~\cite{borth2013large} proposed a concept named Adjective Noun Pairs (ANPs) to simultaneously preserve emotional and location information of objects in an image. 
In order to sum up different level of emotion features, low-level features from elements-of-art, mid-level features from principles-of-art, as well as high-level features from ANPs and facial expressions were combined in a multi-graph learning manner by Zhao~\etal~\cite{zhao2014affective}.
Chen~\etal~\cite{borth2013large} implemented object detection methods to recognize the top six frequent objects (\ie car, dog, dress, face, flower, and food), and proposed a novel classification approach to model the concept similarity between ANPs. 
These hand-crafted features were specifically designed yet limited to represent, which failed to cover all important factors in VEA and resulted in degraded performance on large-scale datasets.

\subsubsection{Deep Learning Methods}
\label{sec:deep_learning_methods}
With the rapid development of deep learning, \textit{Convolutional Neural Networks (CNNs)} were applied in various computer vision tasks and achieved the state-of-the-art performance. 
Due to its great success, researchers in VEA also exploited deep CNN for sentiment prediction. 
Based on their previous work SentiBank~\cite{borth2013large}, Chen~\etal~constructed DeepSentiBank~\cite{chen2014deep} by replacing the SVM classifier with deep CNN, which resulted in a great improvement in both annotation accuracy and retrieval performance. 
You~\etal~\cite{you2015robust} proposed a novel progressive CNN architecture (PCNN) to make use of the noisy labeled data for binary sentiment classification. 
To help training with deep CNN, You~\etal~\cite{you2016building} established a benchmark for emotion recognition by designing a large scale dataset based on the Mikel's eight emotion categories. 
A multi-level deep representation network (MldrNet)~\cite{rao2016learning} was proposed by Rao~\etal~to extract emotion features from multi-scale patches (\ie pixel-level feature, aesthetic feature and semantic feature), which was an early attempt to design an emotion-specific network. 
In order to integrate the features extracted from different levels more effectively, Zhu~\etal~\cite{zhu2017dependency} adopted Bi-GRU as feature fusion module while keep traditional CNN as feature extraction module.
Further, Rao~\etal~\cite{rao2019multi} proposed a multi-level deep network to utilize both low-level and high-level features for predicting visual emotions.
Recently, Yang~\etal~\cite{yang2018visual} constructed a global-local network, implemented object detection method to find affective regions in local branch and combined two branches to make final sentiment prediction.
Yang~\etal~\cite{yang2018weakly} proposed a weakly supervised coupled network (WSCNet) with two branches to leverage the localized information through attention mechanism.
Similarly, He~\etal~\cite{he2019multi} proposed a multi-attentive pyramidal mode, aiming to find emotional cues from local regions and their relationships.
A novel CNN model was proposed by Zhang~\etal~\cite{zhang2019exploring} to extract and integrate content information as well as style information to infer visual emotions.
From a higher perspective, Zhang~\etal~\cite{zhang2020object} propose a novel object semantics sentiment correlation model (OSSCM) to predict emotions from object semantics.
However, most of the existing deep learning methods directly fed the whole image into general deep networks for feature extraction, without a careful consideration of the unique evocation process involved in VEA.

Traditional methods designed specific emotion features manually, which resulted in limited representation ability and degraded performance when facing massive data.
Oppositely, deep learning methods were capable of extracting effective features automatically yet lacked interpretability. 
Learning from both approaches, we first select emotional stimuli manually based on the psychological theories and then implement deep networks to extract emotion features from selected stimuli. 
Based on S-O-R model, we construct a manually stimuli selection and a deep feature extraction network, with both advantages of interpretability and effectiveness. 
Besides, compared with previous methods, we make the first attempt to introduce stimuli selection into VEA, which is proved interpretable and effective.

\begin{figure*}
	\centering
	\includegraphics[width=0.95\textwidth]{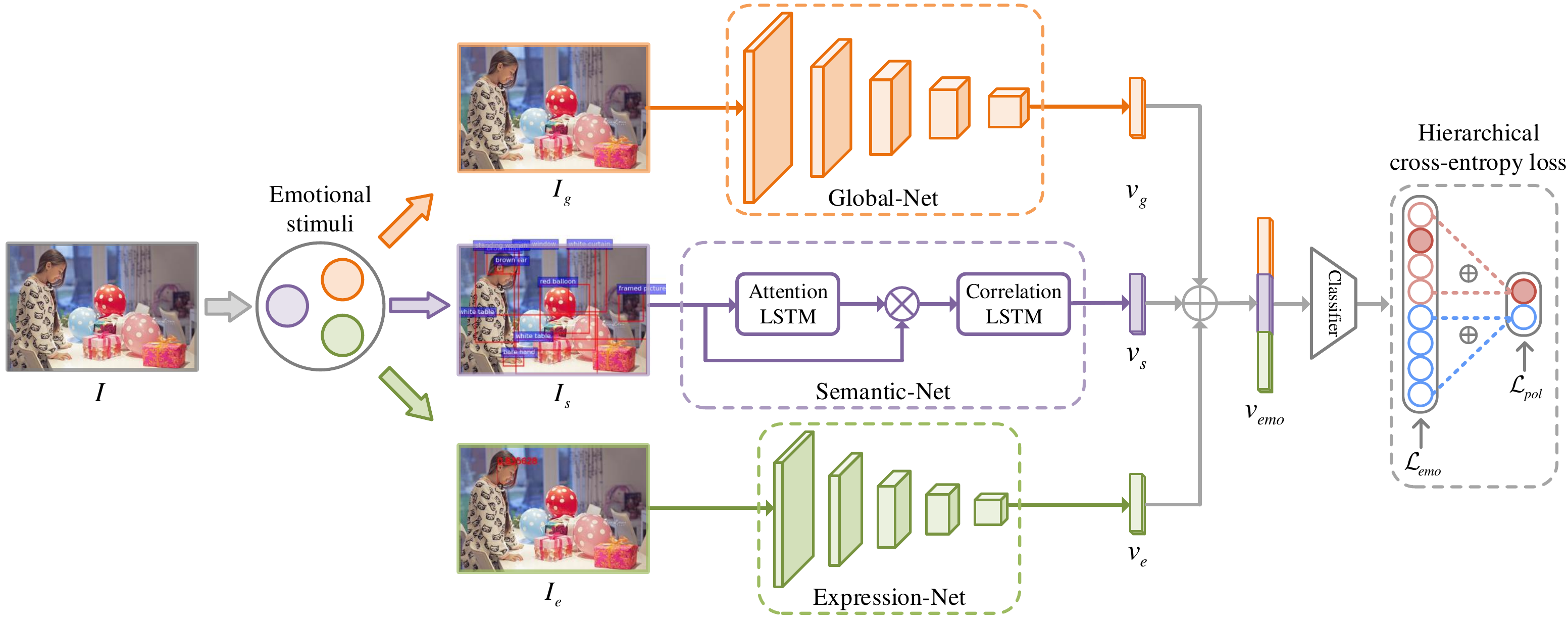}
	\caption{Framework of the proposed stimuli-aware VEA network.
		Based on the S-O-R psychological model, our network consists of three stages, namely stimuli selection, feature extraction and sentiment prediction. 
		First, specific emotional stimuli (\ie color, object, face) are selected from images by employing the off-the-shelf tools in deep learning methods (Sec.~\ref{sec:stimuli_selection}).
		Then, we construct three specific networks (\ie Global-Net, Semantic-Net and Expression-Net) to extract distinct emotion features from different stimuli simultaneously (Sec.~\ref{sec:feature_extraction}).
		Finally, considering the inherent structure of emotion categories in Mikel's wheel, we propose a novel hierarchical cross-entropy loss penalizing hard false examples from easy ones (Sec.~\ref{sec:emotion_prediction}).} 
	\label{fig:network}
\end{figure*}

\subsection{Object Detection}
\label{sec:object_detection}
As one of the basic problems in computer vision, object detection was heated discussed and consequently made great progress recently. 
Due to its high accuracy and efficiency, object detection was implemented into various tasks as a preprocessing stage, such as image caption, VQA, and visual reasoning~\cite{zhou2019re}. Among all the object detection algorithms, several benchmarks were worth-mentioning, including R-CNN~\cite{girshick2014rich}, fast R-CNN~\cite{girshick2015fast}, and faster R-CNN~\cite{ren2015faster}. 
R-CNN, as a pioneering work in this field, implemented deep CNN into object detection for the first time, which greatly improved the detection performance. 
By replacing the original serial structure with a parallel structure, Girshick proposed a new method and named it fast R-CNN, which led to a higher speed as well as improved accuracy. 
The biggest innovation in faster R-CNN was the Region Proposal Network (RPN), as opposed to the selective-search in previous methods. 
RPN not only made it possible to generate effective proposals in an end-to-end network, but it also improved the detection speed to a large extent. 
Considering its great advantage in both speed and accuracy, we employ faster R-CNN as the object detection method in our network.

\subsection{Facial Expression Recognition}
\label{sec:facial_expression_recognition}
\textit{Facial expression recognition (FER)} aims to recognize facial changes in response to a person's internal emotional states, which drawn great interest among researchers and was widely used in real-world applications in recent years ~\cite{xiang2015hierarchical, wang2017regularizing, xiang2018linear, li2018reliable,shojaeilangari2015robust,chiranjeevi2015neutral}.
In FER, it is essential to pre-process the image with face alignment, data augmentation and face normalization~\cite{li2018deep}.
In face alignment, the first step is to detect the face~\cite{li2018deep}.
Dlib~\cite{king2009dlib} and MTCNN~\cite{zhang2016joint} are two of the most commonly-used face detectors in FER.
Considering its wide applications in recent years~\cite{moore2014semantic,lee2018spatiotemporal}, we employ Dlib as our face detector.
After pre-processing, the face image is then sent into a feature extraction network, in which CNN is a common practice. 
Various datasets were collected for facial expression recognition, including lab datasets and web datasets.
Lab datasets consisted of face images specially take in the laboratory environment~\cite{lucey2010extended,pantic2005web,valstar2010induced}, while web datasets were gathered from the internet~\cite{goodfellow2013challenges,fabian2016emotionet,mollahosseini2017affectnet}. 
As most of our emotion datasets were collected from the web, we initialize our Expression-Net with pre-trained parameters from a widely used web dataset named FER2013~\cite{goodfellow2013challenges}.
FER2013 was a large-scale dataset collected automatically on the web by Google image search API, which was introduced during the ICML 2013 Challenges in Representation Learning.

\section{Methodology}
\label{sec:methodology}
In this section, we propose a novel stimuli-aware network for emotion predictions through mining emotions from different emotional stimuli.
Inspired by the S-O-R psychological model, our network is designed with three stages, \ie stimuli selection (S), feature extraction (O) and emotion prediction (R).
To the best of our knowledge, it is the first time to introduce stimuli selection (S) into VEA in an end-to-end network.
Fig.~\ref{fig:network} shows the architecture of the proposed method.
Firstly, specific emotional stimuli (\ie color, object, face) are selected from images by employing the off-the-shelf detection tools in deep learning methods (Sec.~\ref{sec:stimuli_selection}).
Considering the uniqueness of different stimuli, we construct three specific branches, namely Global-Net, Semantic-Net and Expression-Net, to extract distinct emotional features from different stimuli simultaneously (Sec.~\ref{sec:feature_extraction}).
Finally, benefiting from the inherent structure of emotion categories, we design a hierarchical cross-entropy loss to distinguish hard false examples from easy ones with an increased penalty (Sec.~\ref{sec:emotion_prediction}).

\subsection{Stimuli Selection}
\label{sec:stimuli_selection}
\begin{figure*}
	\centering
	\includegraphics[width=0.95\linewidth]{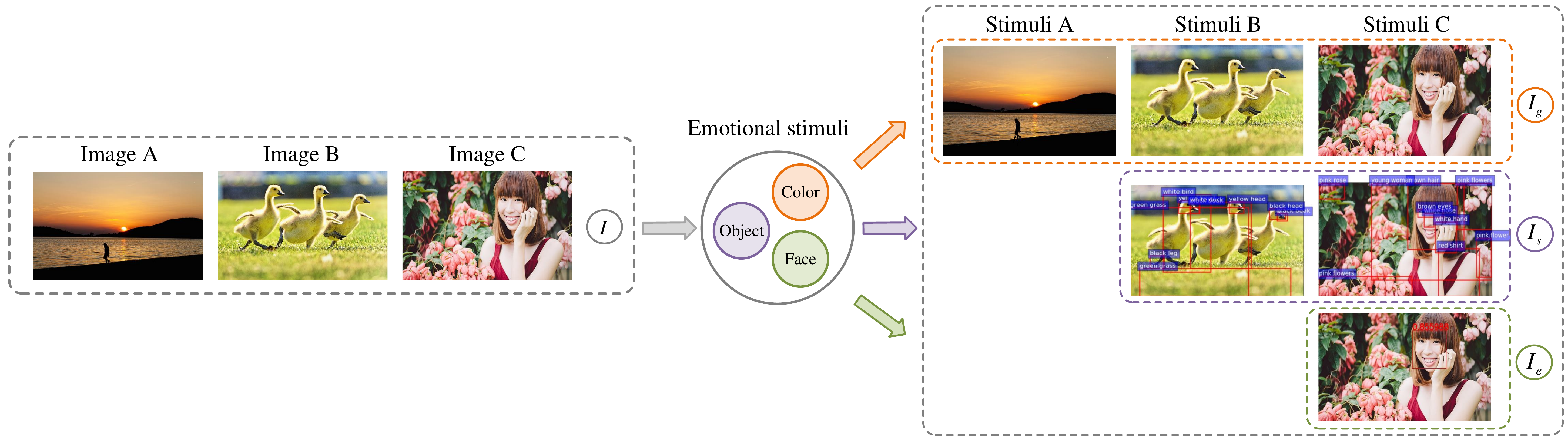}
	\caption{Overview of the stimuli selection process. 
		Based on psychological theory, we choose color, object and face as typical emotional stimuli and implement the off-the-shelf deep learning tools to select these stimuli from our datasets.
		Through this process, an affective image is turned into a set of emotional stimuli, which are specifically designed and further utilized for emotion prediction.}
	\label{fig:metho_1}
\end{figure*}

Unlike other traditional computer vision tasks~\cite{girshick2014rich,girshick2015fast,ren2015faster,krizhevsky2012imagenet,simonyan2014very,he2016deep}, VEA is much more challenging as emotions are complex and ambiguous, which cannot be ``found'' directly in an image.
Apparently, it is easy to classify an image to the category of dog while it is thorny to judge an image with sad.
Therefore, how to effectively bridge the \textit{affective gap}~\cite{zhao2018affective} between low-level pixels and high-level emotions has become the biggest challenge.
Fortunately, great attention has been paid to VEA not only in computer vision but also in psychology.
Psychologist Frijda~\cite{frijda1986emotions} found out that some special substances may evoke emotion and named them emotional stimuli.
Different from the previous stimuli-response (S-R) model, psychologist Woodworth~\cite{woodworth1930dynamic} argued the importance of organism (O).
Based on this, psychologists Mehrabian and Russell~\cite{mehrabian1974approach} further suggested that human perceive emotions through three steps -- stimuli, organism and response, which was the so-called S-O-R model~\cite{buxbaum2016sor, cao2018exploring, luqman2017empirical}.
As mentioned in~\cite{brosch2010perception}, ``How we perceive our environment is thus shaped by categorization processes which guide and constrain the organization of incoming stimulus information,'' indicating the importance of stimuli (S) in S-O-R model.
Further, psychologist Brosch suggested that color, certain objects, facial expressions or any other attributes can be categorized as emotional stimuli~\cite{brosch2010perception}.
Therefore, unlike previous methods, we introduce stimuli selection to VEA, aiming to predict emotions from specific emotional stimuli instead of the whole image.
Inspired by the S-O-R model, we construct our network with three stages, \ie stimuli selection (S), feature extraction (O) and emotion prediction (R), where color, object and face are chosen as typical emotional stimuli according to psychological theories.

We further illustrate the stimuli selection process with three examples in Fig.~\ref{fig:metho_1}. 
Image A is a scenery image describing sunset, with hardly any object or face in it. 
After fed into the stimuli selection module, stimuli A is generated from image A with color stimulus alone.
We can infer from the pale-yellow color stimulus that image A implies \textit{sad} emotion. 
Image B shows three little cute ducks running on the grass. 
Different from the previous scenery image, there are distinct objects in image B, which can be viewed as object stimulus. 
Both object stimulus and color stimulus contribute to the final emotion prediction of \textit{awe}.
Image C describes a smiling girl standing behind a flower-covered wall, which seems comfortable and satisfied. 
Feeding image C into the stimuli selection module and we will get stimuli C containing face stimulus, object stimulus and color stimulus simultaneously. 
Specifically, we can see a happy girl in face stimulus, a wall covered with pink flowers in object stimulus, and the color of the entire picture in color stimulus. 
All the three emotional stimuli are indispensable to the final emotion prediction of \textit{contentment}.
To sum up, different images may have different combinations of emotional stimuli.
Once any of them occurred, it contributes to the final emotion prediction, alone or together with other emotional stimulus.

Therefore, we construct three branches for stimuli selection, \ie color, object and face. 
To be specific, we keep the entire image for color stimulus, \ie ${{I}_{g}=I}$.
Meanwhile, object stimulus (${{I}_{s}}$) and face stimulus (${{I}_{e}}$) are further selected from the entire image by employing the off-the-shelf detection tools in deep learning methods.

For the object branch, we employ faster R-CNN as our stimulus detector. 
To obtain a concrete and precise description of objects, we initialize our faster R-CNN with the pre-trained parameters from Visual Genome dataset~\cite{krishna2017visual}, replacing the original PASCAL VOC~\cite{everingham2007pascal} and MS COCO~\cite{lin2014microsoft} datasets.
Specifically, in the training process, our object detector is only feed forward with a set of frozen parameters.
In other words, our object detector in only a general detection tool which is never specially optimized using emotion datasets alone.
We visualize the object proposals in stimuli selection as is shown in Fig.~\ref{fig:metho_1}.
As discussed above, strong correspondence does not exist between a single object and a certain emotion, which makes it hard to predict the emotion of the whole image from a single object.
Hence, we select multiple objects from an image by ranking their detection confidence.
Specifically, after stimuli selection, image $I$ is turned into object stimulus ${{I}_{s}}$, containing top-$N$ detected objects. We denote the object proposals as ${{I}_{s}}=\left\{ {{i}_{1}},...,{{i}_{N}} \right\}$ and the object features for each proposal as ${{\mathbf{F}}_{s}}=\left\{ {{\mathbf{f}}_{1}},...,{{\mathbf{f}}_{N}} \right\}$, which are automatically extracted from faster-RCNN. 
$N$ denotes the number of regions with $N=10$, which is further ablated with experiments in Sec.~\ref{sec:hyper_parameters_analysis}.
Considering the possible redundancy caused by multiple objects, a specific feature extraction network is further proposed to infer the semantic correlations between different object features ${\mathbf{F}}_{s}$, which will be discussed in Sec.~\ref{sec:semantic_net}.

In order to select face stimulus, we employ landmark face detector Dlib~\cite{king2009dlib} from facial expression recognition. In detail, Dlib first detects faces in each image and then crops them into an aligned size automatically. To be specific, taking the 448$\times$448 image $I$ as input, the face detector will output a face crop ${{I}_{e}}$ with size 48$\times$48, which can be regarded as the face stimulus. 
It is obvious that not all the images in FI dataset contain human faces. If an image does not contain any face, the face detector will output nothing. Besides, when multiple faces appear in an image at the same time, we only choose the largest face as ${{I}_{e}}$, assuming that people are mostly likely to be attracted and thus affected by it.
Through the stimuli selection process, an affective image is turned into a set of emotional stimuli, containing color, object and face, which are specifically designed for further emotion prediction in the following sections.

\subsection{Feature Extraction}
\label{sec:feature_extraction}
After stimuli selection, ${{I}_{g}}$, ${{I}_{s}}$ and ${{I}_{e}}$ are generated as emotional stimuli representing image $I$. Furthermore, we construct three specific branches, namely Global-Net, Semantic-Net and Expression-Net, to extract distinct emotion features from different stimuli simultaneously.

\subsubsection{Global-Net}
\label{sec:global_net}
As mentioned in Sec.~\ref{sec:stimuli_selection}, color, objects, faces or other attributes can be summarized as emotional stimuli. 
Color feature is a global feature showing a strong correlation with emotions. 
However, other global features (\eg texture, pattern, composition, scene) also contribute to the evocation process of emotion and thus cannot be neglected.
Besides, in deep learning methods, there is no specialized network for color feature extraction.
Therefore, we construct Global-Net to extract color feature and other global features.

Most of the existing methods~\cite{yang2018visual,yang2018weakly} in VEA employ Convolutional Neural Networks (CNNs) to extract global features, due to the powerful representation ability of deep features.
In this work, our Global-Net is based on ResNet-50~\cite{he2016deep}, balancing both efficiency and accuracy. 
Specifically, Global-Net consists of five convolutional layers and a Global Averaging Pooling (GAP) layer, which is described in Eq.~\eqref{eq:res50}.
In Global-Net, global stimulus ${{I}_{g}}$ is first sent into the fully convolutional networks of ResNet-50 and then fed into the GAP layer to obtain the global feature vector:
\begin{align}
\label{eq:res50}
{{\mathbf{F}}_{g}}&={\mathrm{FCN}_{\text{res}_{50}}}({{I}_{g}}),\\
{{\textbf{v}}_{g}}&={{G}_{avg}}({{\mathbf{F}}_{g}}),
\end{align}
where $\mathrm{FCN}_{\text{res}_{50}}$ is the fully convolutional networks in ResNet-50 and ${{G}_{avg}}$ is the following GAP layer.
${{{\mathbf{F}}}_{g}}\in {{\mathbb{R}}^{{d_1}\times w\times h}}$ denotes the feature maps extracting from the last convolutional layer and ${{\textbf{v}}_{g}}\in {{\mathbb{R}}^{{d_1}}}$ represents the extracted global vector.
Specifically, $w$, $h$ are the spatial size of the feature map while ${d_1}$ is the length of the global vector with ${d_1}=2048$.


\subsubsection{Semantic-Net}
\label{sec:semantic_net}
\begin{figure}
	\centering
	\includegraphics[width=0.95\linewidth]{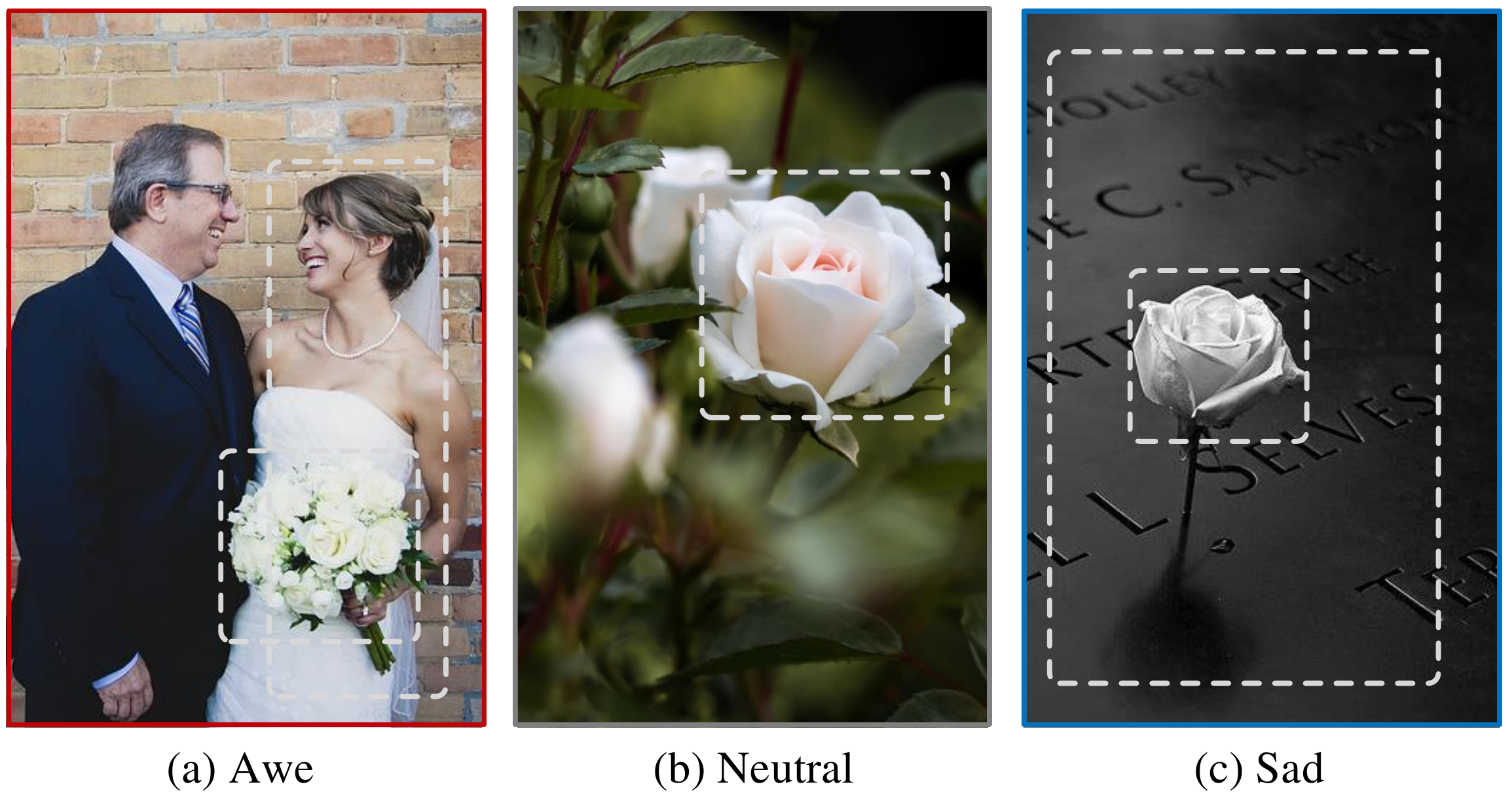}
	\caption{Examples from FI dataset (\ie (a), (c)) and the internet (\ie (b)).
		It is not the white rose alone that evokes a specific emotion, but the interactions between different objects jointly determine the final result.}
	\label{fig:metho_2}
\end{figure}

In Sec.~\ref{sec:stimuli_selection}, we implement object detection methods to select multiple objects in an image by ranking their detection confidence. 
Due to the complexity in emotion evocation process, it is not always an isolated object itself evokes a specific emotion, but the semantic correlations between different objects jointly determine the final result. For example, when we see white roses in Fig.~\ref{fig:metho_2}(b), it is hard to decide whether a positive emotion or a negative one.
However, as shown in Fig.~\ref{fig:metho_2}(a), when white roses are accompanied by a bride, we can infer that it is a wedding and are very likely to evoke a positive emotion of \textit{awe}.
Oppositely, when a white rose and a tombstone come together in Fig.~\ref{fig:metho_2}(c), we may have a negative feeling of \textit{sad}.
Therefore, instead of using the multi-object features directly, it is more reasonable to mine the semantic correlations between different objects.

Researchers have recently drawn great attention on exploiting the relationships between different objects, including Visual Question Answering (VQA), image captioning, visual reasoning, \textit{etc.} 
In image caption, a newly released up-down attention model~\cite{anderson2018bottom} has achieved great performance by implementing attention mechanism to weigh different object features. 
Based on this model, we design a emotion-specific Semantic-Net to mine the semantic correlations between different objects in our network.
In order to exploit the correlations between different objects, we treat object features as a sequential data, and propose a Long-Short Term Memory (LSTM)~\cite{hochreiter1997long} network to model such dependencies.
The LSTM network is composed of two LSTM layers, namely attention LSTM and correlation LSTM, which is a standard implementation~\cite{donahue2015long}.
Firstly, we design an attention LSTM aiming to weigh each object features as well as reducing the redundancy of multiple object features.
After that, correlation LSTM is constructed to mine the semantic correlations between different objects.

\begin{figure*}
	\centering
	\includegraphics[width=0.75\linewidth]{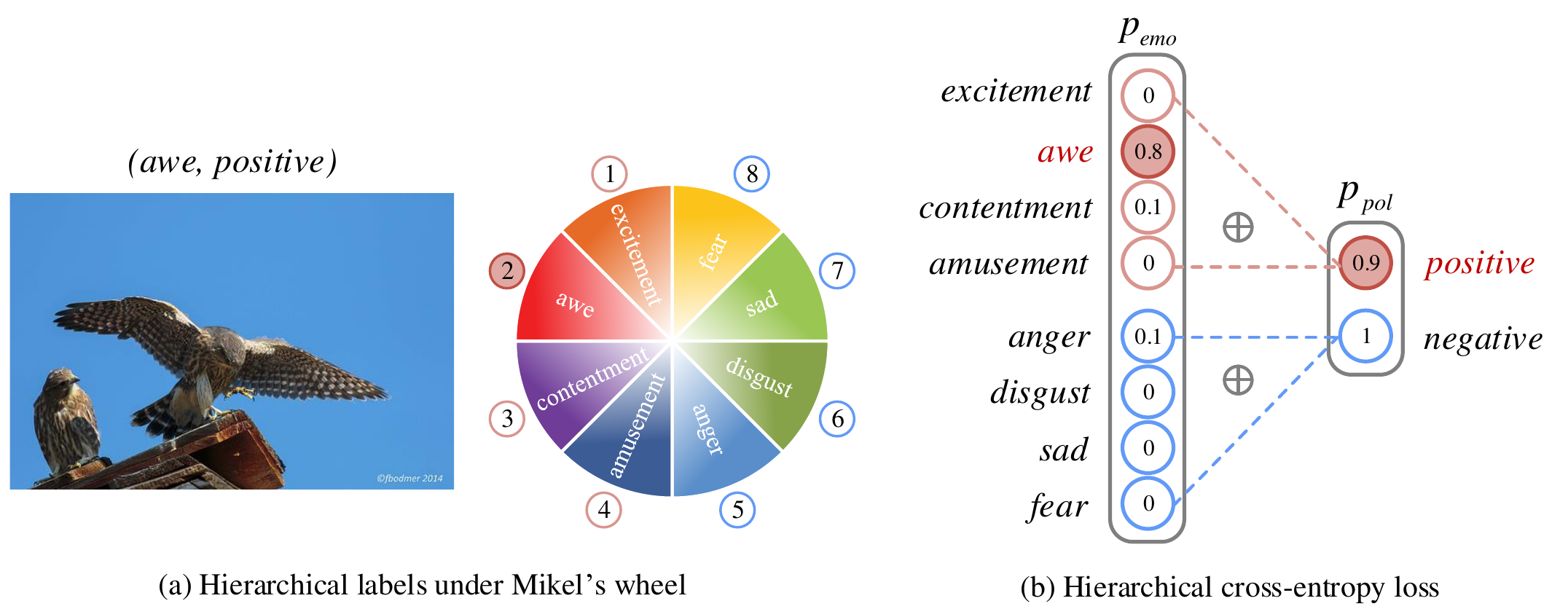}
	\caption{Detailed illustration of the proposed hierarchical cross-entropy loss.
		Our hierarchical cross-entropy is motivated by the inherent structure of Mikel's wheel from psychology (a) and is further calculated corresponding to such hierarchical structure (b).}
	\label{fig:loss_2}
\end{figure*}
After object stimulus selection, we obtain object proposals ${{I}_{s}}=\left\{ {{i}_{1}},...,{{i}_{N}} \right\}$ and corresponding object features ${{\mathbf{F}}_{s}}=\left\{ {{\mathbf{f}}_{1}},...,{{\mathbf{f}}_{N}} \right\}$, in which $N$ represents the number of objects in an image. In attention LSTM, $\mathbf{x}_{t}^{att}$ denotes the input vector while $\mathbf{h}_{t}^{att}$ denotes the output vector. The input vector $\mathbf{x}_{t}^{att}$ to the attention LSTM at each time step consists of the previous hidden state  $\mathbf{h}_{t-1}^{cor}$ of the correlation LSTM, concatenated with the mean-pooled object features ${{\mathbf{f}}_{mean}}=\frac{1}{N}\sum\limits_{i=1}^{N}{{{\mathbf{f}}_{i}}}$.
The whole process of attention LSTM is given by
\begin{align}
\label{eq:h_att}
\mathbf{h}_{t}^{att}&=\mathrm{{LSTM}_{att}}\left( \mathbf{x}_{t}^{att},\mathbf{h}_{t-1}^{att} \right),\\
\label{eq:x_att}
\mathbf{x}_{t}^{att}&=\left[ \mathbf{h}_{t-1}^{cor},{{\mathbf{f}}_{mean}} \right].
\end{align}
The output $\mathbf{h}_{t}^{att}$ of attention LSTM is then fed into Bahdanau attention~\cite{bahdanau2014neural} module, together with object features ${{\mathbf{f}}_{i}}$. The attention module generates a normalized attention weight ${{\alpha }_{i,t}}$ for each object features ${{\mathbf{f}}_{i}}$ at time step $t$ as
\begin{align}
\label{eq:a}
{{a}_{i,t}}&={{\omega }_{a}}\tanh \left( {{\textbf{W}}_{f}}{{\textbf{f}}_{i}}+{{\textbf{W}}_{h}}\textbf{h}_{t}^{att} \right),\\
\label{eq:alpha}
{{\alpha }_{i,t}}&=\mathrm{softmax} \left( {{a}_{i,t}} \right),
\end{align}
where ${{\textbf{W}}_{f}}\in {{\mathbb{R}}^{M\times F}}$ , ${{\textbf{W}}_{h}}\in {{\mathbb{R}}^{M\times H}}$ and ${{\omega }_{a}}\in {{\mathbb{R}}^{M}}$ are learned parameters of the attention module.
The input object features ${{\textbf{F}}_{s}}=\left\{ {{\textbf{f}}_{1}},...,{{\textbf{f}}_{N}} \right\}$ are then weighed by the attention coefficient ${{\alpha }_{i,t}}$. The weighted object features, as shown in Eq.~\eqref{eq:att}, are then fed into correlation LSTM:
\begin{align}
\label{eq:att}
{{\textbf{f}}_{att}}&=\sum\limits_{i=1}^{N}{{{\alpha }_{i,t}}{{\textbf{f}}_{i}}}.
\end{align}
In correlation LSTM, $\textbf{x}_{t}^{cor}$ denotes the input vector while $\textbf{h}_{t}^{cor}$ denotes the output vector. The input vector is concatenated by previous hidden state $\textbf{h}_{t-1}^{att}$ of the attention LSTM and the weighted object features ${{\textbf{f}}_{att}}$, given by
\begin{align}
\label{eq:h_cor}
\textbf{h}_{t}^{cor}&=\mathrm{{LSTM}_{cor}}\left( \textbf{x}_{t}^{cor},\textbf{h}_{t-1}^{cor} \right),\\
\label{eq:x_cor}
\textbf{x}_{t}^{cor}&=\left[ \textbf{h}_{t-1}^{att},{{\textbf{f}}_{att}} \right].
\end{align}
After conducting both attention LSTM and correlation LSTM, not only has the redundancy between multiple objects been reduced, but the correlation between different objects has also been mined.
Therefore, the output of the Semantic-Net can be viewed as semantic feature with a higher-level representation of objects, which is given as
\begin{align}
\label{eq:v_s}
 {{\textbf{v}}_{s}}=\textbf{h}_{T}^{cor},
\end{align}
where $T$ denotes the last time step in LSTM and ${{\textbf{v}}_{s}}\in {{\mathbb{R}}^{{d_2}}}$ represents the extracted semantic feature with ${d_2}=512$.

\subsubsection{Expression-Net}
\label{sec:expression_net}

It is obvious that human facial expression is the most direct way to convey emotions. In our daily life, if you see someone with a certain emotion, you will probably be affected by the emotion and evoke a similar emotion as well. This is called empathy, which was introduced by psychologist Jim Rogers~\cite{rogers1975empathic}. Empathy is the ability to experience the inner world of other people, which is a peculiar ability of human beings. Therefore, facial expression is considered as an indispensable factor in VEA, which is designed as a branch in our network as well.

Since face images are small in size, previous researchers usually implement a shallow deep network to extract facial expression features. In light of this, we employ ResNet-18 as the base net to construct Expression-Net.
Our Expression-Net is first initialized with pre-trained parameters from FER2013 dataset~\cite{goodfellow2013challenges} and then jointly fine-tuned on FI dataset~\cite{you2016building} with the rest networks.
Expression-Net consists of five convolutional layers and a GAP layer, which is similar to Global-Net.
It is worth mentioning that not all the images in our affective datasets have face stimulus.
Therefore, the design of Expression-Net is divided into two cases as shown in Eq.~\eqref{eq:v_e}.
If face stimulus ${{I}_{e}}$ exists, we first send it into the fully convolutional networks of ResNet-18 and then feed it into the GAP layer to obtain the expression feature ${{\textbf{v}}_{e}}$.
If face detector cannot find any face stimulus, we directly set ${{\textbf{v}}_{e}}$ to a zero vector, which does not contribute to the emotion prediction:
\begin{align}
\label{eq:v_e}
{{\textbf{v}}_{e}}=\left\{ \begin{array}{*{35}{l}}
\!\!\!{{G}_{avg}}(\mathrm{{FCN}_{res_{18}}}({{I}_{e}})),{{\textbf{v}}_{e}}\in {{\mathbb{R}}^{{d_3}}}\!\!\!\!\!\!\!& ,\!\text{ }{{I}_{e}}\text{ exsits }  \\
\!\!\!\mathbf{0} & ,\!\text{ else }  \\
\end{array} \right.
\end{align}
where $\mathrm{FCN}_{\text{res}_{18}}$ is the fully convolutional networks in ResNet-18 and ${{G}_{avg}}$ is the following GAP layer.
Besides, ${{\textbf{v}}_{e}}\in {{\mathbb{R}}^{{d_3}}}$ represents the extracted expression feature with ${d_3}=512$.

\subsection{Emotion Prediction}
\label{sec:emotion_prediction}
\subsubsection{Feature Fusion}
\label{sec:feature_fusion}
In the previous sections, we select three typical emotional stimuli (Sec.~\ref{sec:stimuli_selection}) and design specific networks to extract distinct features from them (Sec.~\ref{sec:feature_extraction}).
As all the three features, \ie global feature (${{\textbf{v}}_{g}}$), semantic feature (${{\textbf{v}}_{s}}$), and expression feature (${{\textbf{v}}_{e}}$), are indispensable and complementary in emotion evocation process, we further concatenate them as
\begin{align}
\label{eq:v_emo}
{{\textbf{v}}_{emo}}=concate\left[ {{\textbf{v}}_{g}},{{\textbf{v}}_{s}},{{\textbf{v}}_{e}} \right],
\end{align}
where ${{\textbf{v}}_{emo}}\in {{\mathbb{R}}^{{d_1}+{d_2}+{d_3}}}$ denotes emotion feature.
The emotion feature ${{\textbf{v}}_{emo}}$ is further used for emotion prediction in Eq.~\eqref{eq:p8}.

\subsubsection{Hierarchical Cross-Entropy Loss}
\label{sec:hierarchical_cross_entropy_loss}

In traditional classification tasks, cross-entropy (CE) loss has been widely used and has achieved great performance recently, where categories are often spatially separated and irrelevant to each other.
For example, in ImageNet~\cite{deng2009imagenet}, cats and pens are two distinct categories, sharing few relationships with each other.
However, unlike other classification tasks, emotions are strongly correlated and share a unique hierarchical structure with each other according to psychological theories~\cite{mikels2005emotional, zhao2016predicting}.

In our work, we implement Mikel's wheel~\cite{mikels2005emotional, zhao2016predicting} as base model with eight emotions, \ie amusement, anger, awe, contentment, disgust, excitement, fear, sad.
In addition to its given category, each emotion is assigned with a specific polarity according to Mikel's model~\cite{mikels2005emotional, zhao2016predicting}.
Specifically, amusement, awe, contentment, excitement are positive emotions, while anger, disgust, fear, sad belong to the negative side.
Based on the above facts, we design a novel hierarchical cross-entropy loss for VEA, aiming to exploit more emotion-specific information from psychological models.
To be specific, we add an auxiliary polarity loss to the traditional cross-entropy loss, with increased penalties on hard false predictions compared with easy ones.

\begin{figure}
	\centering
	\includegraphics[width=0.55\linewidth]{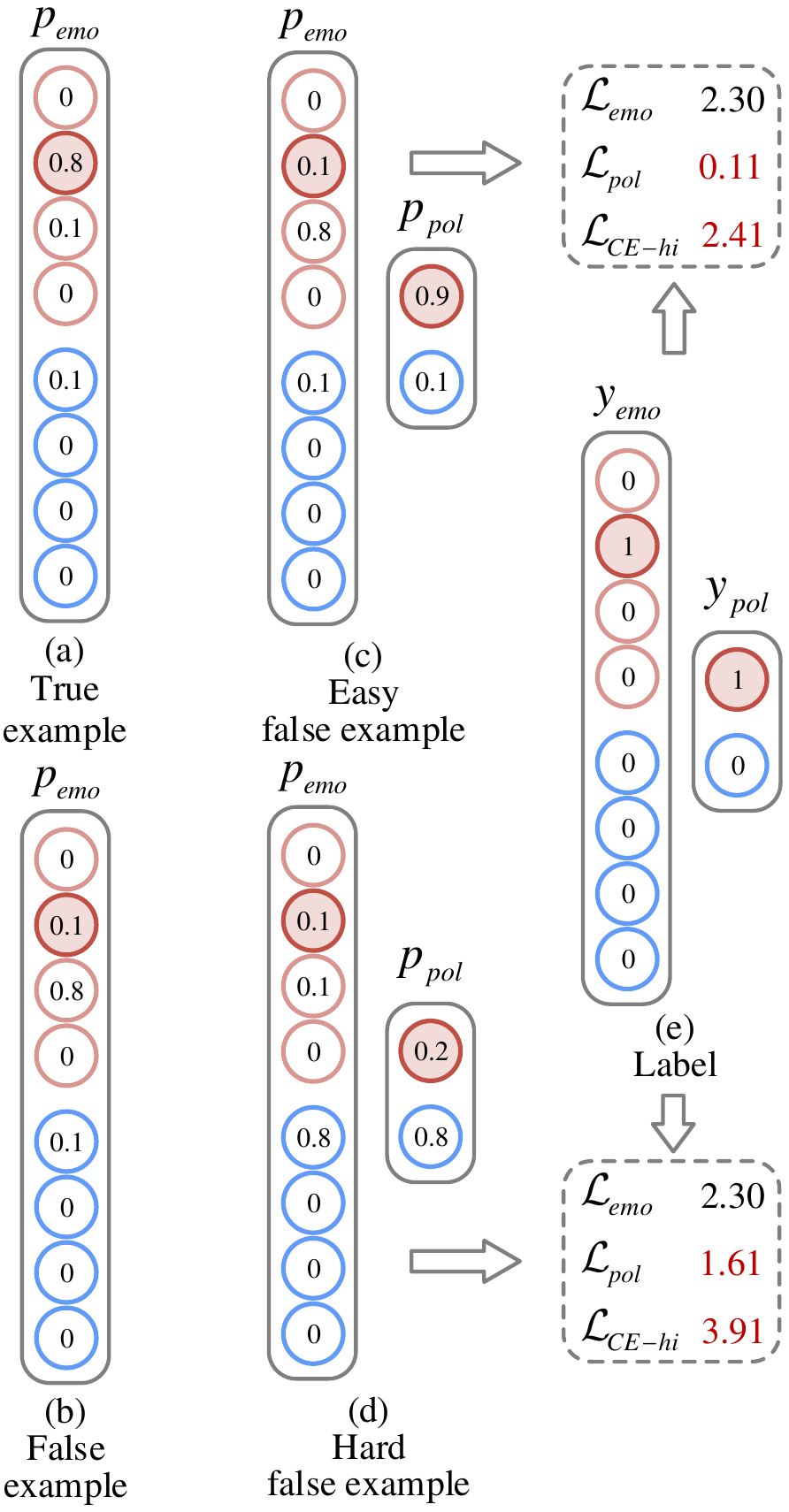}
	\caption{Numerical examples to validate the effectiveness of the proposed hierarchical cross-entropy loss.
		It is suggested that emotion loss alone cannot distinguish hard false example from the easy one while the proposed loss is able to separate them apart with an increased penalty.}
	\label{fig:loss_3}
\end{figure}
We first send the concatenated emotion feature ${{\textbf{v}}_{emo}}$ to a classifier and a softmax function successively:
\begin{align}
\label{eq:p8}
{{p}_{emo}}\left( i\left| {{\textbf{v}}_{emo}},\textbf{W} \right. \right)=\frac{\exp \left( {{\textbf{w}}_{i}}{{\textbf{v}}_{emo}} \right)}{\sum_{i=1}^{C}{\exp \left( {{\textbf{w}}_{i}}{{\textbf{v}}_{emo}} \right)}},
\end{align}
where ${{p}_{emo}}\in{\mathbb{R}}^{C}$ represents emotion vector and $C$ denotes the number of emotion categories in our datasets.
$\textbf{W}\in {{\mathbb{R}}^{({d_1}+{d_2}+{d_3})\times C}}$ is a learnable weight matrix of the emotion classifier, which consists of ${\textbf{w}}_{i}$.

Since ${{p}_{emo}}$ is a probability vector, we can easily obtain $\sum\nolimits_{i=1}^{C}{{{p}_{emo}}\left( i \right)=1}$, where each position in ${{p}_{emo}}(i)$ corresponds to the predicted probability of a specific emotion $i$ in Mikel's model, as shown in Fig.~\ref{fig:loss_2}.
In Fig.~\ref{fig:loss_2} (a), take an affective image as an example, we represent the hierarchical labels, \ie emotion label and polarity label, under the inherent structure of Mikel's wheel.
To be specific, the positive emotions (\ie excitement, amusement, contentment, awe) are correspond to the former $C/2$ positions in ${{p}_{emo}}$, while the negative (\ie sad, fear, disgust, anger) the rest $C/2$ respectively.
Based on the above partition, our emotion vector ${{p}_{emo}}$ can be correspondingly summed up to a polar vector ${{p}_{pol}}$ in accordance with Mikel's model:
\begin{align}
\label{eq:positive}
positive:{{p}_{pol}}\left( 0 \right)&=\sum\limits_{i=1}^{C/2}{{{p}_{emo}}\left( i \right)},\\
\label{eq:negative}
negative:{{p}_{pol}}\left( 1 \right)&=\!\!\sum\limits_{i=C/2}^{C}\!\!{{{p}_{emo}}\left( i \right)},
\end{align}
where polar vector is also a probability vector that satisfies $\sum\nolimits_{j=0}^{1}{{{p}_{pol}}\left( j \right)=1}$.
Each value in ${{p}_{pol}}$ can also be viewed as the prediction probability of a specific polarity, \ie positive, negative.
Besides the formulas in Eq.~\ref{eq:positive}--\ref{eq:negative}, we also illustrate the partition process in Fig.~\ref{fig:loss_2} (b) for better comprehension.

In traditional cross-entropy loss, there are only two kinds of predictions: true examples {(\ie Fig.~\ref{fig:loss_3} (a)) and false ones (\ie Fig.~\ref{fig:loss_3} (b)).
However, considering the unique hierarchical structure in emotion classification, false predictions can be divided into simple false examples (\ie Fig.~\ref{fig:loss_3} (c)) and hard false ones (\ie Fig.~\ref{fig:loss_3} (d)).
As shown in Fig.~\ref{fig:loss_3}, we define simple false examples with emotions incorrectly classified and polarities correctly classified. 
Similarly, hard false refers to examples whose both emotion and polarity are incorrectly classified.
In order to distinguish hard false examples from easy ones, we proposed an auxiliary polarity loss with increased penalties on those hard ones, which is shown in Eq.~\ref{eq:pol_l}.
Meanwhile, true examples and false ones are separated by implementing traditional cross-entropy loss on emotion label, \ie emotion loss, as shown in Eq.~\ref{eq:emo_l}:
\begin{align}
\label{eq:pol_l}
{{\mathcal{L}}_{pol}}=\sum\limits_{j=0}^{1}{{{y}_{pol}}\left( j \right)\log \left( {{p}_{pol}}\left( j \right) \right)},\\
\label{eq:emo_l}
{{\mathcal{L}}_{emo}}=\sum\limits_{i=1}^{C}{{{y}_{emo}}\left( i \right)\log \left( {{p}_{emo}}\left( i \right) \right)},
\end{align}
where ${y}_{pol}$ in Eq.~\ref{eq:pol_l} represents the polarity label and ${y}_{emo}$ in Eq.~\ref{eq:emo_l} denotes the emotion label in our datasets.

To illustrate the necessity and effectiveness of the proposed hierarchical cross-entropy loss, we cite a simple example in Fig.~\ref{fig:loss_3}.
Compare Fig.~\ref{fig:loss_3} (c) with (e), we can see that with emotion loss alone, the easy false example is treated as identical as the hard one, \ie with a same numerical result at $2.30$.
However, polarity loss of two examples are distinguishable from each other, \ie with a huge discrepancy between $0.11$ and $1.61$.
Thus, the auxiliary polarity loss helps the network to distinguish easy false examples and hard ones apart with increased penalties on hard ones, which is a fine-grained constraint towards precise emotion prediction.
Both losses, \ie emotion loss and polarity loss, are essential and indispensable to the final emotion prediction, which are further combined to form the hierarchical cross-entropy loss:
\begin{align}
\label{eq:ce_hi}
{\mathcal{L}_{CE\text{-}hi}}&={{\mathcal{L}}_{emo}}+\lambda {{\mathcal{L}}_{pol}},
\end{align}
where $\lambda$ is a hyper-parameter balancing the importance between the two losses and is further ablated in Sec.~\ref{sec:ablation_study}.
Finally, the whole network is optimized through Eq.~\ref{eq:ce_hi} jointly.
Therefore, the proposed hierarchical cross-entropy loss can effectively guide the network to pay more attention to hard false examples and consequently helps the network to predict emotions in a hierarchical and emotion-specific manner.

\section{Experimental results}
\label{sec:experimental_results}
In this section, we first evaluate our approach on four widely-used datasets, including FI~\cite{you2016building}, EmotionROI~\cite{peng2016emotions}, Artphoto~\cite{machajdik2010affective}, IAPSa~\cite{lang1999international,mikels2005emotional}, compared with the state-of-the-art methods. 
Moreover, ablation study is conducted to further validate the effectiveness of our network.

\subsection{Datasets}
\label{sec:datasets}
\textbf{FI dataset.} The FI dataset~\cite{you2016building} is currently the largest well-labeled dataset, containing approximately 23,000 images, which are selected from the Flickr and Instagram by searching eight emotion categories as keywords, \ie amusement, anger, awe, contentment, disgust, excitement, fear and sad. The weakly labeled emotional images are then well-labeled by 225 Amazon Mechanical Turk (AMT) workers. At least receiving three agreements among five workers, the weakly labels and their corresponding images are preserved as the final FI dataset.

\textbf{EmotionROI.} The EmotionROI dataset~\cite{peng2016emotions} is a sentiment prediction benchmark collected from Flickr, which contains 1980 images with six emotion categories, \ie anger, disgust, fear, joy, sad, surprise. Besides categorical label, each image is also annotated on pixel-level with 15 responses corresponding to different emotional regions.

\textbf{ArtPhoto.} Images in ArtPhoto dataset~\cite{machajdik2010affective} are taken by professional artists, aiming to evoke a certain emotion in their photos. There are in total 806 images in this dataset and each of them is labeled with a specific emotion from the eight emotion categories.

\textbf{IAPSa dataset.} The International Affective Picture System (IAPS) dataset~\cite{lang1999international} is a normative emotional stimuli dataset for visual sentiment analysis research. The IAPSa dataset~\cite{mikels2005emotional} is a subset of the IAPS dataset, which contains 395 images and is labeled with the Mikel's eight sentiment categories.

The above four datasets can be categorized as large-scale dataset (\ie FI dataset) and small-scale ones (\ie EmotionROI, ArtPhoto, IAPSa dataset).

\begin{figure}
	\centering
	\includegraphics[width=\linewidth]{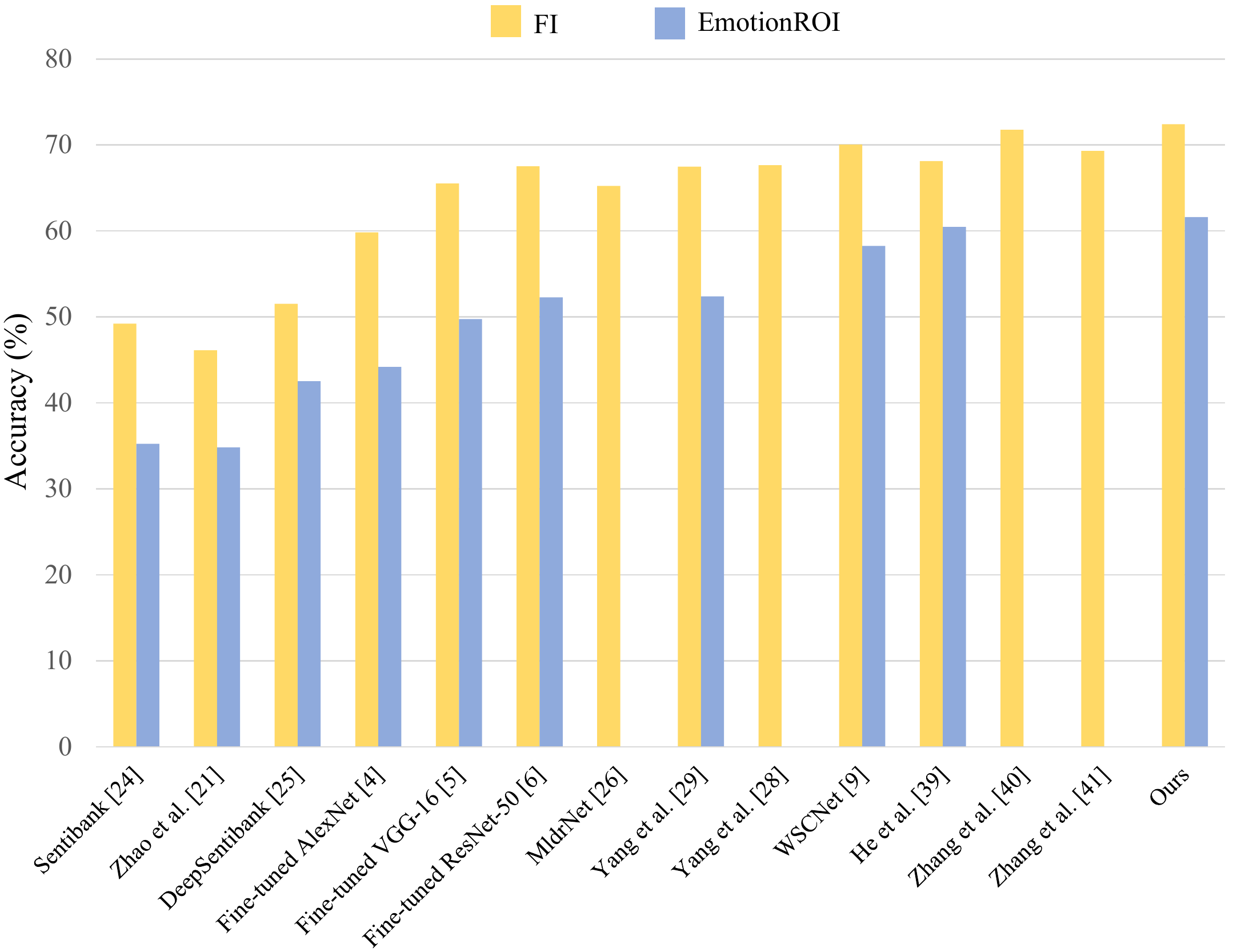}
	\caption{Classification results on FI dataset and EmotionROI dataset, compared with the state-of-the-art methods.}
	\label{fig:chart_sota}
\end{figure}
\begin{table}
	\centering
	\caption{Classification accuracy (\%) on FI dataset and EmotionROI dataset, compared with the state-of-the-art methods.}
	\label{tab:FI_dataset}
	\renewcommand\arraystretch{1.18}
	\begin{tabular}{ccc}
		\toprule
		\toprule
		Method & FI & EmotionROI\\
		\midrule
		Sentibank~\cite{borth2013large} & 49.23 & 35.24\\
		Zhao~\etal~\cite{zhao2014exploring} & 46.13 & 34.84\\
		DeepSentibank~\cite{chen2014deep} & 51.54 & 42.53\\
		Fine-tuned AlexNet~\cite{krizhevsky2012imagenet} & 59.85 & 44.19\\
		Fine-tuned VGG-16~\cite{simonyan2014very} & 65.52 & 49.75\\
		Fine-tuned ResNet-50~\cite{he2016deep} & 67.53 & 52.27 \\
		MldrNet~\cite{rao2016learning} & 65.23 & --\\
		Yang~\etal~\cite{yang2017joint} & 67.48 & 52.40\\
		Yang~\etal~\cite{yang2018retrieving} & 67.64 & --\\
		WSCNet~\cite{yang2018weakly} & 70.07 & 58.25\\
		He~\etal~\cite{he2019multi} & 68.13 & 60.47\\
		Zhang~\etal~\cite{zhang2019exploring} & 71.77 & --\\
		Zhang~\etal~\cite{zhang2020object} & 69.32 & --\\
		Ours & \textbf{72.42} & \textbf{61.62}\\
		\bottomrule
		\bottomrule
	\end{tabular}
\end{table}

\subsection{Implementation Details}
\label{sec:implementation_details]}
To extract specific emotion features, we initialize our Global-Net with pre-trained parameters from ImageNet~\cite{deng2009imagenet} while Expression-Net from FER2013~\cite{goodfellow2013challenges}. 
Besides, our object stimuli detector is initialized with pre-trained parameters from Visual Genome dataset~\cite{krishna2017visual}, aiming to obtain an accurate description on objects. 
The whole network is then jointly fine-tuned on emotion datasets with the proposed hierarchical cross-entropy loss in an end-to-end manner.

\begin{figure*}
	\centering
	\includegraphics[width=0.9\linewidth]{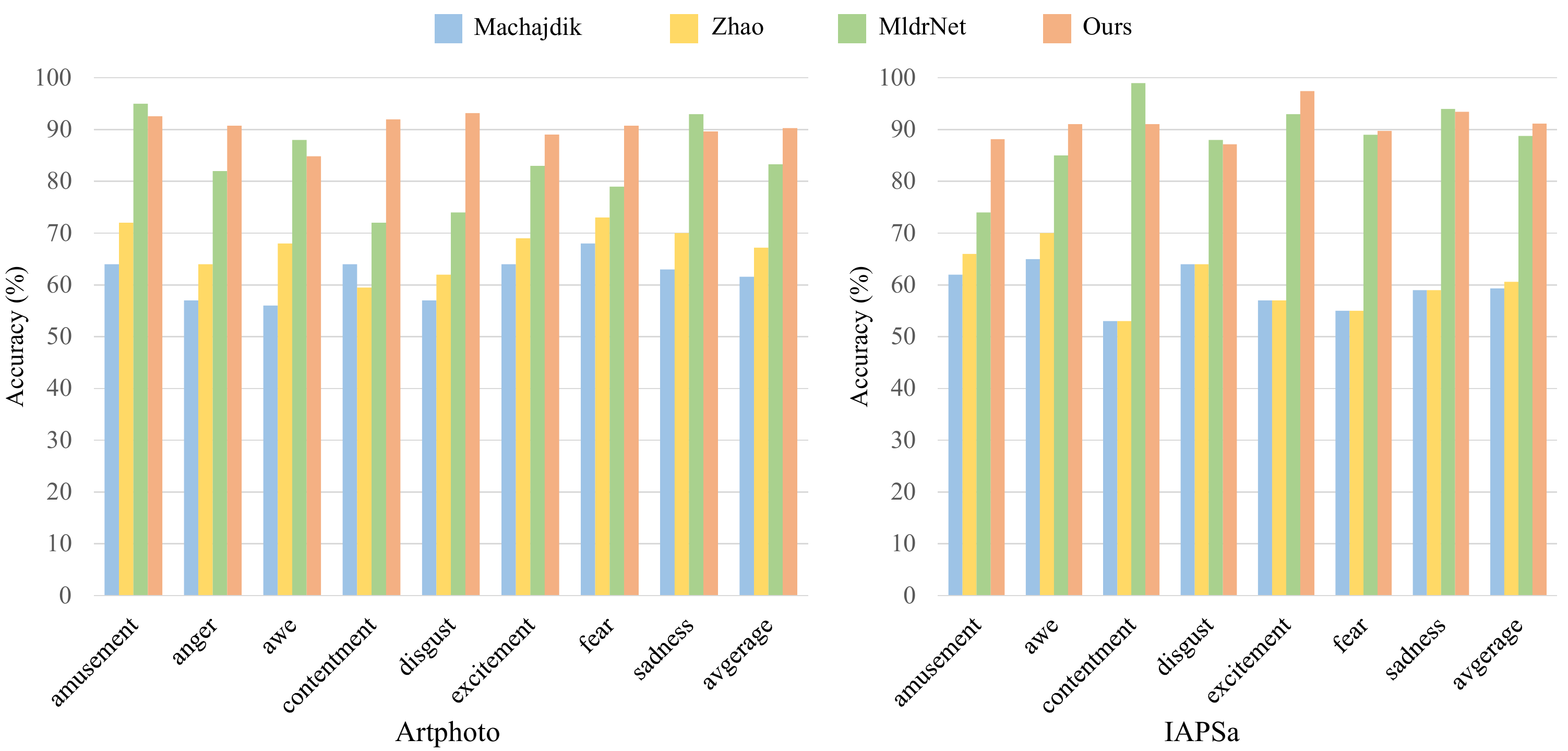}
	\caption{Classification results on two small-scale datasets, \ie Artphoto and IAPSa dataset, compared with the state-of-the-art methods.}
	\label{fig:chart_iapsa_art}
\end{figure*}

The FI dataset is randomly split into training set (80\%), validation set (5\%) and testing set (15\%), following the same setting in~\cite{you2016building}.
The small-scale datasets are split into training set (80\%) and testing set (20\%) randomly except the one with specified training/testing split, \ie EmotionROI~\cite{peng2016emotions}, following the previous work~\cite{yang2017joint, yang2018weakly, he2019multi}.
In addition, each image is resized with its shorter side to 480 and then cropped to 448$\times$448 randomly from the original image or its horizontal flip~\cite{he2016deep}. 
We train all our models by using the adaptive optimizer Adam~\cite{kingma2014adam}, with a weight decay of 5e-5. 
The learning rate starts from 5e-5 and is decayed by 0.1 every 5 epochs, and the total epoch number is set to 50. 
Our framework is implemented using PyTorch~\cite{paszke2017automatic} and our experiments are performed on an NVIDIA GTX 1080ti GPU.

\begin{table}
	\centering
	\caption{Ablation study of network structure on FI dataset.}
	\label{tab:ablation}
	\renewcommand\arraystretch{1.18}
	\begin{tabular}{cccccc}
		\toprule
		\toprule
		\multicolumn{2}{c}{G-Net}&\multicolumn{2}{c}{S-Net}&{\multirow{2}{*}{E-Net}}& {\multirow{2}{*}{Acc (\%)}}\\
		\cmidrule(lr){1-2}
		\cmidrule(lr){3-4}
		{RGB} & {Y}  &LSTMs & FC layers&&\\
		\midrule
		&$\checkmark$&&&& 64.16\\
		$\checkmark$&&&&& 67.93\\
		&&$\checkmark$&&& 66.47\\
		&&&&$\checkmark$& 31.64\\
		$\checkmark$&&$\checkmark$&&& 70.93\\
		$\checkmark$&&&&$\checkmark$& 70.81\\
		&&$\checkmark$&&$\checkmark$& 68.06\\
		&$\checkmark$&$\checkmark$&&$\checkmark$& 69.69\\
		$\checkmark$&&&$\checkmark$&$\checkmark$& 71.39\\
		$\checkmark$&&$\checkmark$&&$\checkmark$& \textbf{72.42}\\
		\bottomrule
		\bottomrule
	\end{tabular}
\end{table}

\subsection{Comparison with the State-of-the-art Methods}
\label{sec:comparison_to_state_of_the_art}
\subsubsection{Comparison on large-scale dataset}
\label{sec:comparison_on_large_scale_dataset}
To evaluate the effectiveness of the proposed method, we first compare our method with the state-of-the-art methods on large-scale dataset FI~\cite{you2016building}, as shown in TABLE~\ref{tab:FI_dataset}. Sentibank~\cite{borth2013large} and Zhao~\etal~\cite{zhao2014exploring} adopted hand-crafted features, which were early attempts in VEA.
Besides, we also conduct experiments on those widely used CNN backbones, including AlexNet~\cite{krizhevsky2012imagenet}, VGG16~\cite{simonyan2014very} and ResNet50~\cite{he2016deep}, which are initialized with the pre-trained parameters on ImageNet~\cite{deng2009imagenet} and then fine-tuned on FI. 
It is obvious that deep learning methods performed favorably against the traditional ones, which attributed to the powerful representation ability of deep features. 
Based on CNN backbones, researchers subsequently developed specific deep networks to deal with visual emotion analysis, including MldrNet~\cite{rao2016learning}, WSCNet~\cite{yang2018weakly}, Zhang~\etal~\cite{zhang2019exploring}, Zhang~\etal~\cite{zhang2020object}, \textit{etc.}
We further show the holistic curve of classification accuracy in Fig.~\ref{fig:chart_sota} to summarize the overall performance of the proposed method. 
Above all, the proposed method outperforms the state-of-the-art methods on FI dataset.

\begin{figure}
	\centering
	\includegraphics[width=0.85\linewidth]{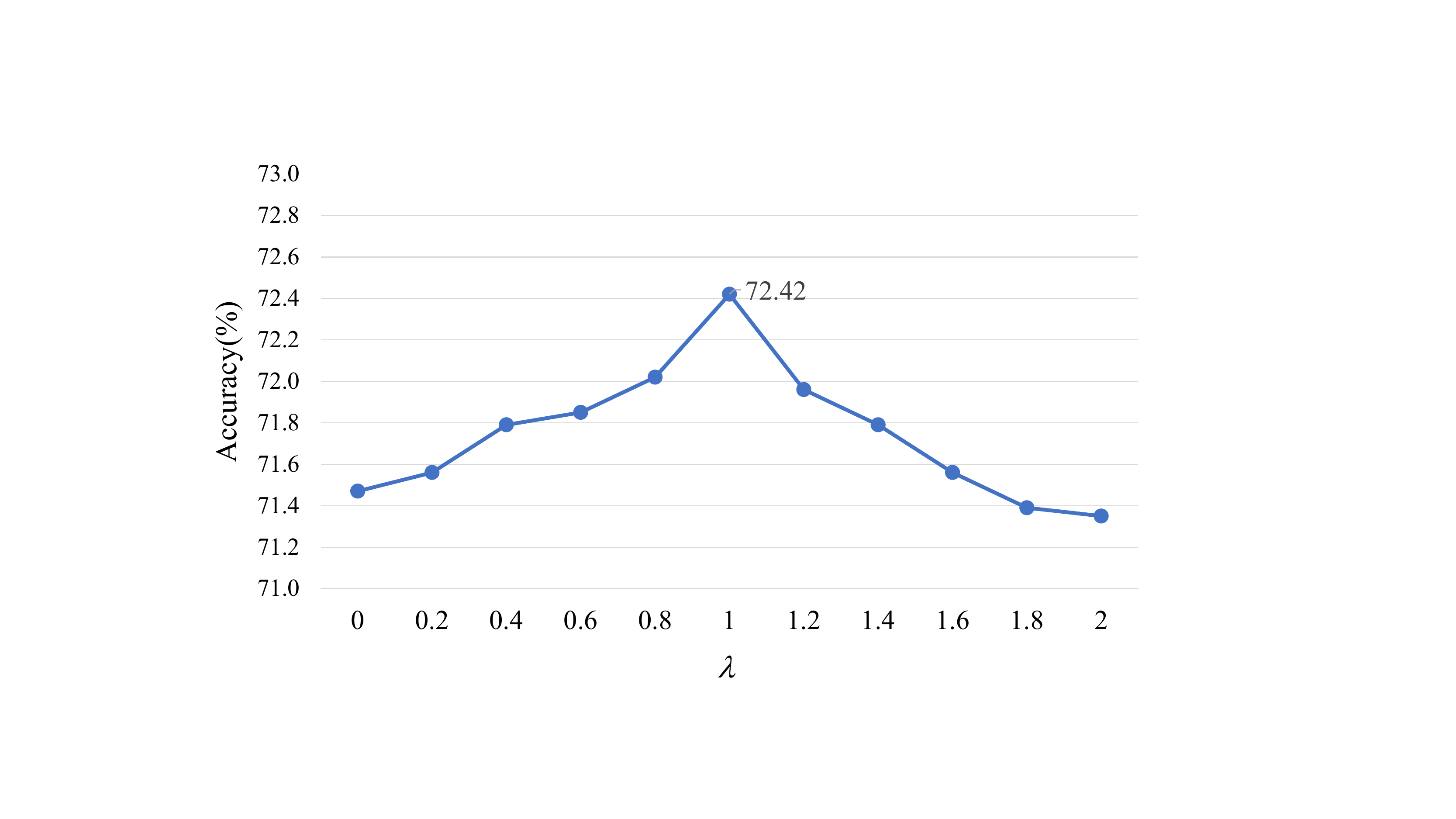}
	\caption{Hyper-parameter analysis of $\lambda$ in the hierarchical cross-entropy loss.}
	\label{fig:chart_lambda}
\end{figure}

\subsubsection{Comparison on small-scale datasets}
\label{sec:comparison_on_small_scale_datasets}
We further verify the effectiveness of the proposed method on three small-scale datasets.
Considering the class-imbalanced and limited images in IAPSa\cite{lang1999international} and Artphoto\cite{machajdik2010affective} dataset, we employ the ``one against all'' strategy to train the classifiers for fair comparison following the previous methods\cite{machajdik2010affective}. 
Besides, image samples in each category are randomly split into five batches and a 5-fold cross-validation is further implemented for the classification results. 
Since the emotion category of anger only contains eight samples, we remove this category for its sample insufficiency following\cite{machajdik2010affective,zhao2014exploring,rao2016learning}.
As shown in Fig.~\ref{fig:chart_iapsa_art}, our method outperforms the state-of-the-art methods, including Machajdik\cite{machajdik2010affective}, Zhao~\etal\cite{zhao2014exploring} and MldrNet~\cite{rao2016learning}.
In addition, we also implement our methods on EmotionROI dataset and validate its superiority towards the state-of-the-art methods, as shown in TABLE~\ref{tab:FI_dataset} and Fig.~\ref{fig:chart_sota}.
It is worth mentioning that the missing data in TABLE~\ref{tab:FI_dataset} and Fig.~\ref{fig:chart_sota}. is caused by lacking of both classification results and open source codes.
Above all, the proposed method consistently outperforms the state-of-the-art methods on both large-scale dataset and small ones, which further proves the effectiveness and robustness of our method.

\begin{figure}
	\centering
	\includegraphics[width=0.85\linewidth]{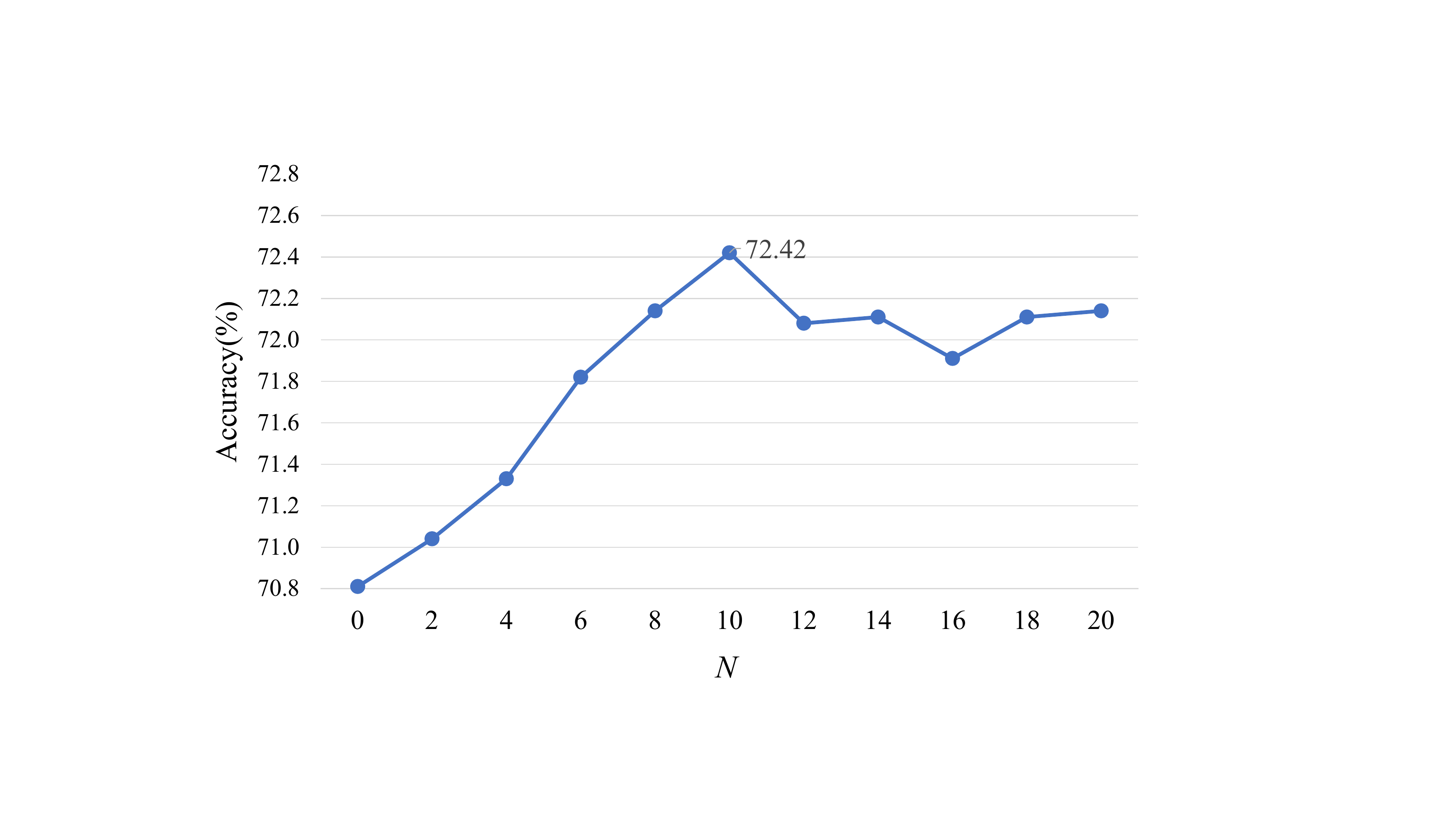}
	\caption{Hyper-parameter analysis of $N$ in the stimuli selection process.}
	\label{fig:chart_N}
\end{figure}

\begin{figure*}
	\includegraphics[width=\linewidth]{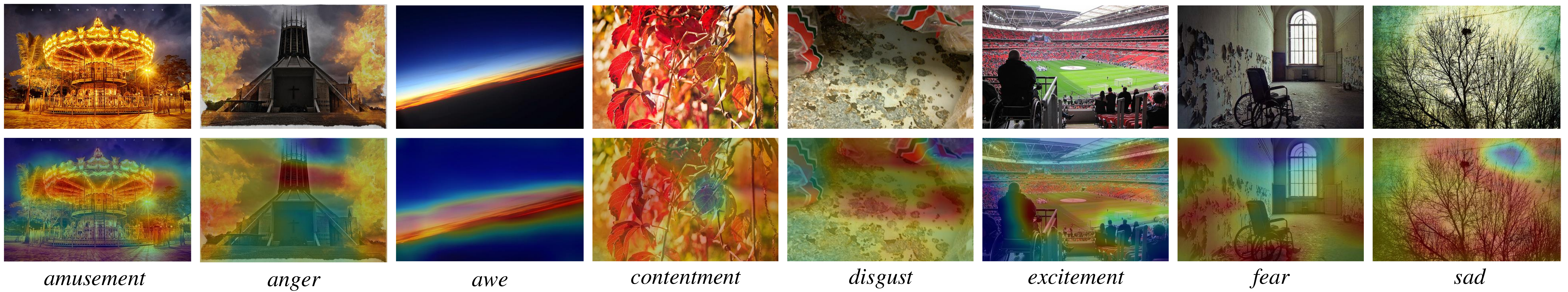}
	\caption{Visualization of Global-Net. It is shown that Global-Net tend to pay more attention to color features.}
	\label{fig:visual_g_8}
\end{figure*}

\begin{figure*}
	\centering
	\includegraphics[width=\linewidth]{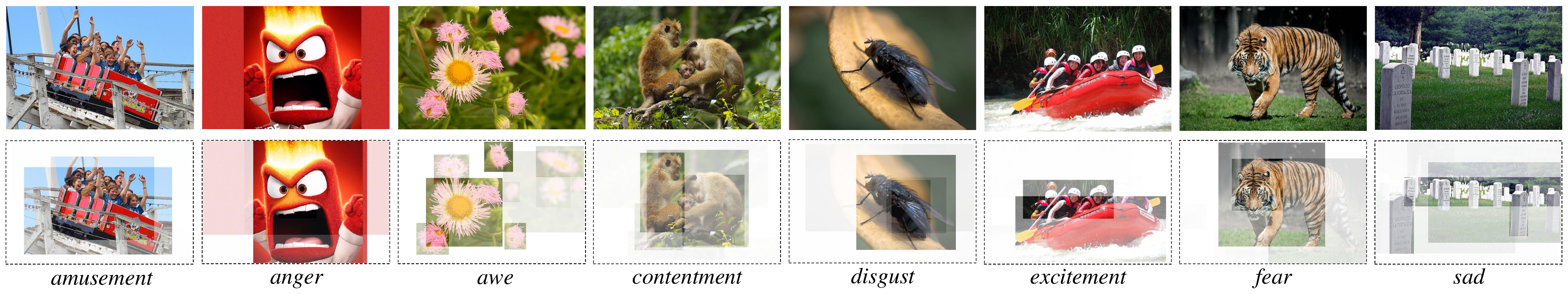}
	\caption{Visualization of Semantic-Net. Semantic-Net extracts the correlations between different objects, which contributes to the final emotion.}
	\label{fig:visual_s_8}
\end{figure*}

\begin{figure*}
	\centering
	\includegraphics[width=\linewidth]{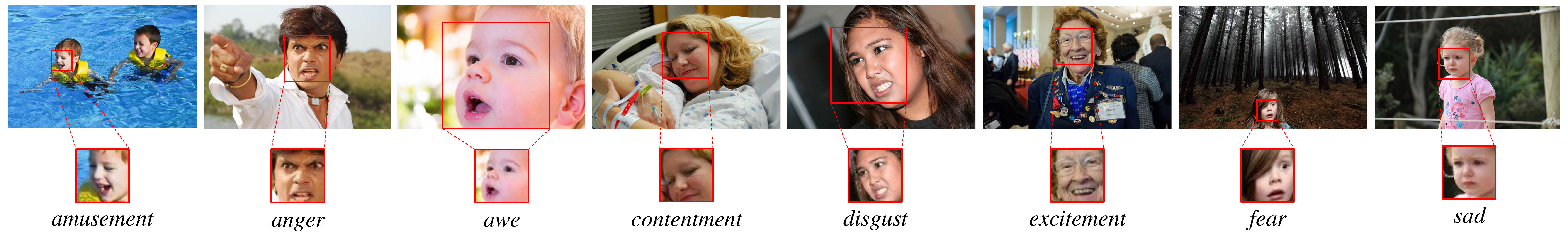}
	\caption{Visualization of Expression-Net. Expressions in affective images greatly impact the viewer's emotions, which is called empathy.}
	\label{fig:visual_e_8}
\end{figure*}

\subsection{Ablation Study}
\label{sec:ablation_study}
\subsubsection{Network architecture analysis}
\label{sec:network_architecture_analysis}
In TABLE~\ref{tab:ablation}, we conduct a set of ablation studies to verify the necessity of each stimulus as well as the effectiveness of each module in the proposed network. 
Our network is divided into three sub-networks, namely Global-Net (G-Net), Semantic-Net (S-Net) and Expression-Net (E-Net).
TABLE~\ref{tab:ablation} shows that when three sub-network acts alone, G-Net reaches the best result, which attributes to the great representation ability of global features. 
Since not every image in FI dataset contains human faces, the poor performance of E-Net alone is explainable and thus we choose it as an auxiliary branch in the whole network.

To validate the necessity of color stimulus, we remove the color information from an image by introducing YCbCr space and restoring Y channel alone.
It is obvious that Y channel alone (\ie 64.16\%) has a great performance drop comparing to RGB channels (\ie 67.93\%), which suggests that color features are essential to emotion prediction and G-Net is capable to extract such information.
Notably, without color features, the three-branch network (\ie 69.69\%) only outperforms the two-branch network (\ie 68.06\%) to a small extent, where both differ from the final result (\ie 72.42\%) to a large extent.
To exploit semantic correlations between different objects, we design two structures for S-Net, namely LSTMs and FC layers, where experimental results show that LSTMs is more capable of extracting semantic correlations. 
In Sec.~\ref{sec:semantic_net}, we demonstrate the superiority of the LSTMs structure theoretically and in this section we prove its effectiveness empirically. 
We can conclude from TABLE~\ref{tab:ablation} that each network structure is indispensable and complementary, which jointly contributes to the final result.

\subsubsection{Hyper-parameters analysis}
\label{sec:hyper_parameters_analysis}
Parameter $\lambda$ controls the proportion of the two elements in the loss function Eq.~\eqref{eq:ce_hi}, which is a decisive hyper-parameter for the classification results. As shown in Fig.~\ref{fig:chart_lambda}, the classification accuracy increases as $\lambda$ grows from 0 to 1 and decreases as $\lambda$ drops from 1 to 2, in which the peak accuracy reaches 72.42\% when $\lambda=1$. When $\lambda$ is set to 0, the hierarchical cross-entropy loss degrades to a general cross-entropy loss, which fails to depict the unique hierarchical structure of emotion categories and leads to a sub-optimal result. As $\lambda$ grows larger, polarity loss dominates the emotion loss, rendering to a decrease in accuracy as well. Hence, we fix $\lambda=1$ in the following experimental settings.

The number of bounding boxes, \ie $N$, is another important hyper-parameters in our method.
Similar to its common setting, we vary $N$ from 0 to 20, as shown in Fig.~\ref{fig:chart_N}.
As $N$ grows from 0 to 10, the classification accuracy is constantly growing. 
In Sec.~\ref{sec:semantic_net}, we assumed that a specific emotion is not evoked by a single objects, but is jointly determined by multiple objects and the correlations between them, which can be further proved by this experiment.
However, there exists a slightly drop after 10, due to information redundancy of too many detection boxes.
From the above analysis, we choose N = 10 as the node number in our stimuli selection process.

\begin{figure}
	\centering
	\includegraphics[width=0.8\linewidth]{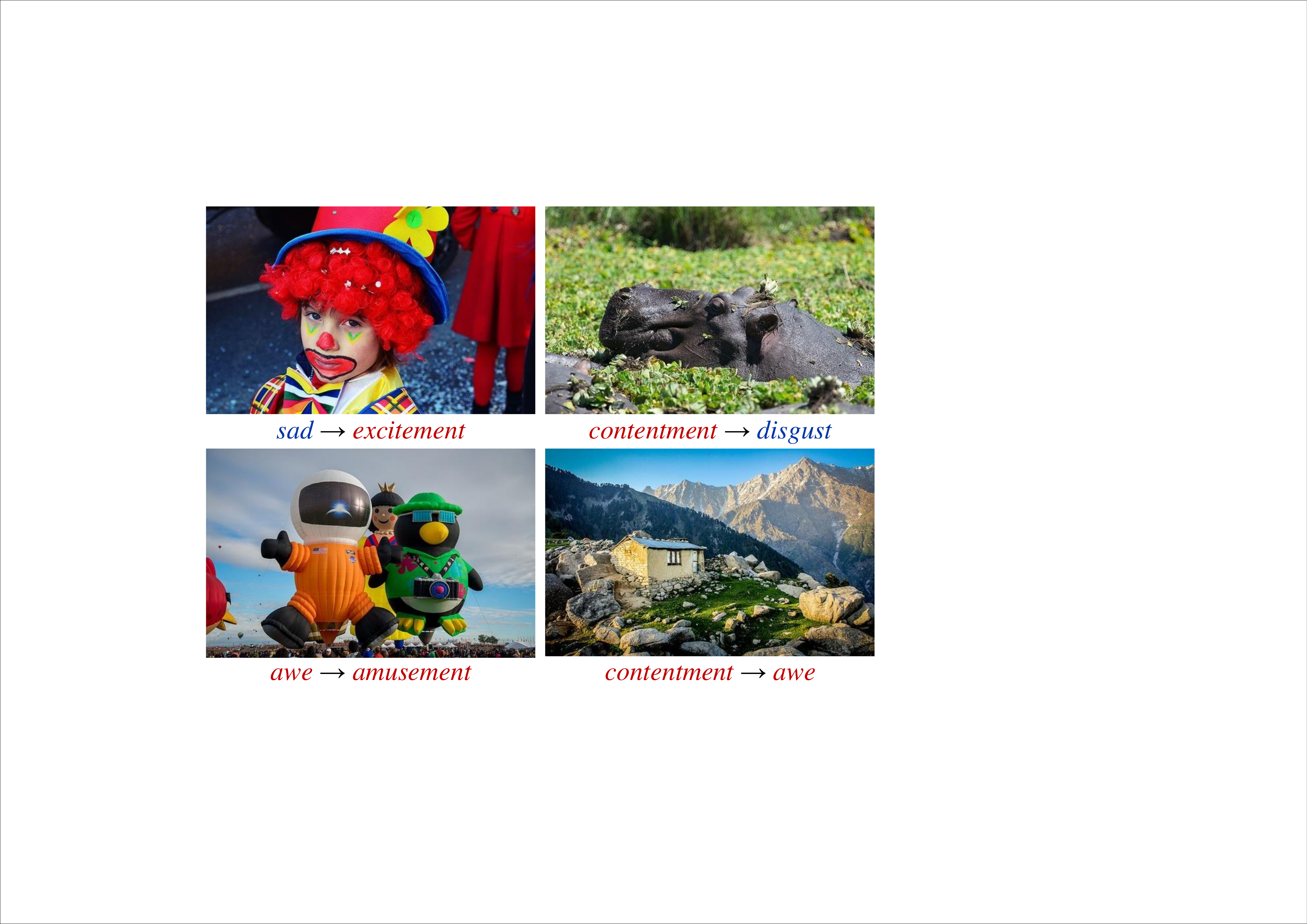}
	\caption{Failure cases on FI dataset, where red denotes positive emotions and blue represents the negative ones.
	The former emotion is the labeled one while the latter the incorrectly predicted one.}
	\label{fig:fail}
\end{figure}

\section{Visualization Results}
\label{sec:visualization_results}
To further prove the effectiveness and interpretability of our network, we visualize the three sub-networks (\ie Global-Net, Semantic-Net, Expression-Net) with examples representing eight different emotions in FI dataset~\cite{you2016building}.
Besides, we also present some failure cases in our method and analyze the potential reasons behind them.

In Global-Net, we directly extract Class Activation Map (CAM) following~\cite{zhou2016learning}, where red represents the most attentive part while blue the least one.
From Fig.~\ref{fig:visual_g_8}, we can infer that Global-Net mainly focuses on global features, especially the color feature, when it comes to an image without any distinct objects or faces.
Take amusement as an example, Global-Net focuses on the warm lights of carousel, which brings us happiness and joy. The example of fear shows an abandoned hospital, suggesting horror and fear from the color of the whole image.

In Semantic-Net, instead of focusing on a single object, we generate multiple object bounding boxes and leverage attention LSTM and correlation LSTM to weigh the importance and mine the correlations between different objects respectively in Sec.~\ref{sec:semantic_net}. In Fig.\ref{fig:visual_s_8}, different bounding boxes represent distinct objects, in which the depth of color indicates the importance of the specific object. First, we can infer from Fig.\ref{fig:visual_s_8} that multiple objects jointly determine the final emotion, as opposed to a single one. Besides, different bounding boxes are correlated, where the bounding box with a deeper color suggests an increased attention weight. Fig.\ref{fig:visual_s_8} shows that it is not a single flower but the flower sea amazes us with awe. In the example of excitement, Semantic-Net not only detects happy people, but also the drift, as drifting evokes great excitement indeed.

In Expression-Net, we visualize the cropped faces directly, since facial expressions is the most obvious stimulus to convey emotions, which is called empathy~\cite{rogers1975empathic}. As mentioned in Sec.~\ref{sec:stimuli_selection}, all the faces are cropped to an aligned size of 48$\times$48, shown in the second row of Fig.\ref{fig:visual_e_8}. From the visualization results, it is obvious that we are easily affected by the expression given in an image and thus Expression-Net is indispensable in our network. Take contentment as an example, there is a new-born baby lying with his mother, where the satisfied mother may affect us with contentment. Expression-Net detects a crying girl from the example of sad, bringing us sad emotion oppositely.

However, there still exist some failure cases in our method.
As shown in Fig.~\ref{fig:fail}, we select some incorrectly classified images from FI dataset, where the former emotion represents the labeled one and the latter the predicted one.
Take the first image as an example, the clown boy wears colorful clothes and made up for an exaggerated smile.
Considering these two emotional stimuli, the network tends to predict it with positive emotion, \ie \textit{excitement}.
However, in human cognition, clowns are often disguised to be happy, especially there exists sadness in the boy's eyes, which involves a high-level common sense in human cognition process.
Therefore, when encountering complex scenarios, it is hard for deep network to learn such prior knowledge through a simple training process. 
Besides, some of the incorrect predictions may be attributed to the noisy labels in our datasets, as shown in the below two images in Fig.~\ref{fig:fail}.
In FI dataset, most images in \textit{amusement} seem like the former one and most images in \textit{awe} look like the latter one, which makes it understandable to make such false predictions.
Above all, failure cases are mainly caused by the complexity and ambiguity of emotions, which may be partly overcome by designing emotion-specific network and labeling datasets by psychologists.

\section{Conclusion}
\label{sec:conclusion}
Inspired by S-O-R model in psychology, we propose a stimuli-aware VEA network to simulate the evocation process of human emotions, which consists of stimuli selection, feature extraction and emotion prediction.
In the first stage, we introduced stimuli selection into VEA for the first time, by implementing the off-the-shelf tools to choose specific emotional stimuli.
After that, we construct three specific sub-networks to extract distinct emotion features from different stimuli simultaneously.
Finally, a hierarchical cross-entropy loss is proposed to distinguish hard false examples from easy ones, which helps the network to learn in an emotion-specific manner.
Experiments demonstrated that the proposed method consistently outperforms the state-of-the-art methods on four widely-used affective datasets. 
Ablation study and visualization results further validated the effectiveness and interpretability of the proposed method.



\bibliographystyle{IEEEtran}
\bibliography{IEEEfull,mybibfile}
%

%

%
%
%




\begin{IEEEbiography}[{\includegraphics[width=1in,height=1.25in,clip,keepaspectratio]{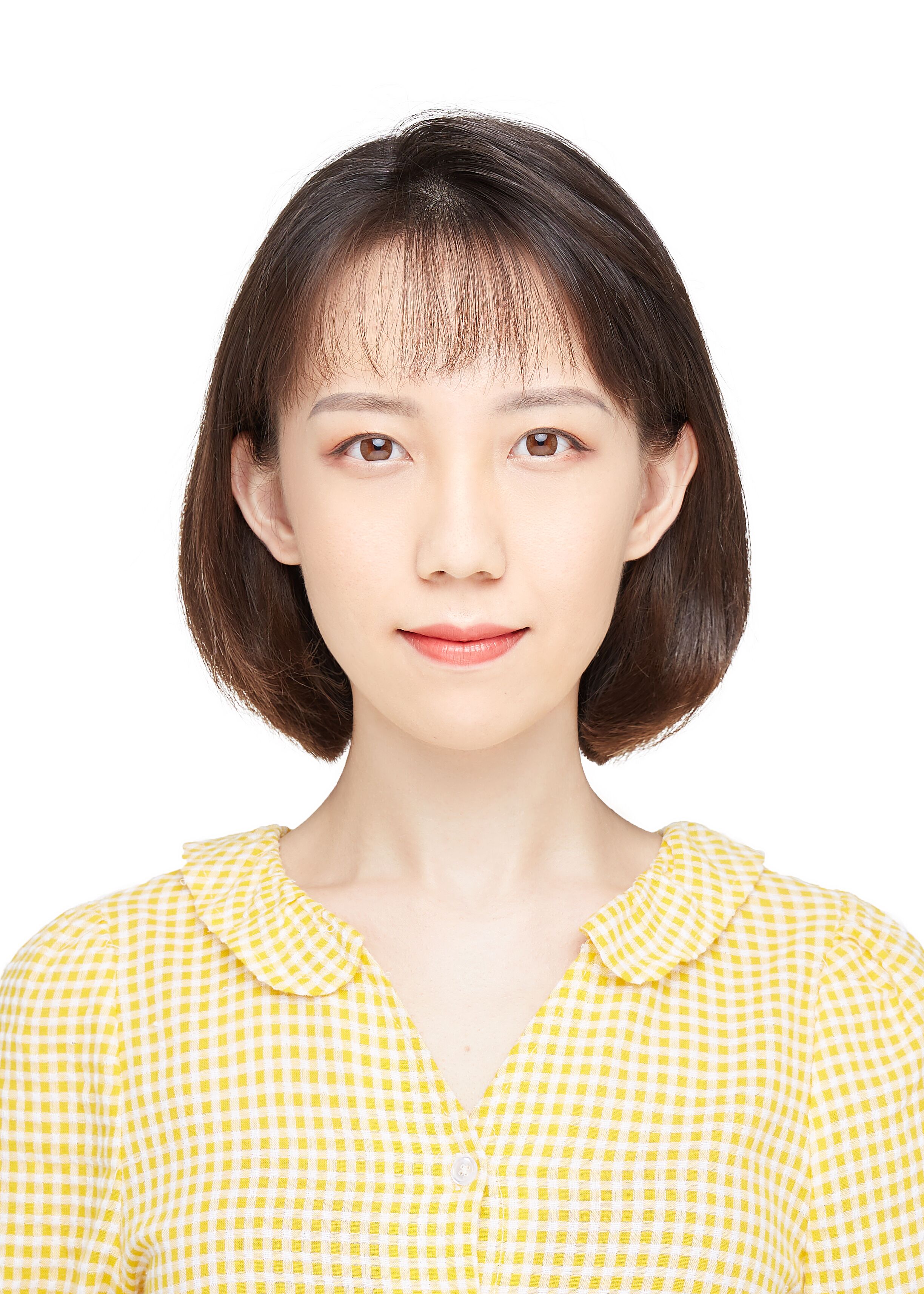}}]{Jingyuan Yang}
	received the B.Eng. degree in Electronic and Information Engineering from Xidian University, Xi'an, China, in 2017. She is currently a Ph. D. Candidate at the School of Electronic Engineering, Xidian University. Her current research interest is visual emotion analysis in deep learning and its applications.
\end{IEEEbiography}

\begin{IEEEbiography}[{\includegraphics[width=1in,height=1.25in,clip,keepaspectratio]{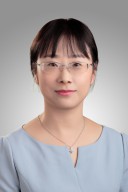}}]{Jie Li}
	received the B.Sc. degree in electronic engi-neering, the M.Sc. degree in signal and information processing, and the Ph.D. degree in circuit and sys-tems from Xidian University, Xi’an, China, in 1995, 1998, and 2004, respectively. She is currently a Professor with the School of Electronic Engineering, Xidian University. Her research interests include image processing and machine learning. In these areas, she has published around 50 technical articles in refereed journals and proceedings, including the IEEE T-IP, T-CSVT, and Information Sciences.
\end{IEEEbiography}

\begin{IEEEbiography}[{\includegraphics[width=1in,height=1.25in,clip,keepaspectratio]{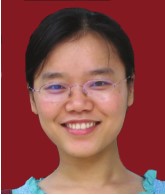}}]{Xiumei Wang}
	received the Ph.D. degree from Xidian University, Xi’an, China, in 2010. She is currently a Lecturer with the School of Electronic Engineering, Xidian University. Her current research interests include nonparametric statistical models and machine learning. She has published several scientific articles, including the IEEE Trans. Cybernetics, Pattern Recognition, and Neurocomputing in the above areas.
\end{IEEEbiography}

\begin{IEEEbiography}[{\includegraphics[width=1in,height=1.25in,clip,keepaspectratio]{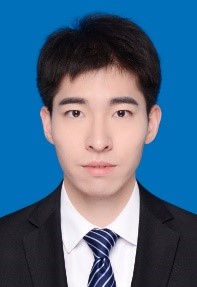}}]{Yuxuan Ding}
	was born in 1995. He received the B.Eng. degree in Intelligent Science and Technology from Xidian University, Xi’an, China, in 2018. He is currently a Ph. D. Candidate at the School of Electronic Engineering, Xidian University. His main research interest covers Machine Learning, Computer Vision, Vision-Language, and their applications.
\end{IEEEbiography}

\begin{IEEEbiography}[{\includegraphics[width=1in,height=1.25in,clip,keepaspectratio]{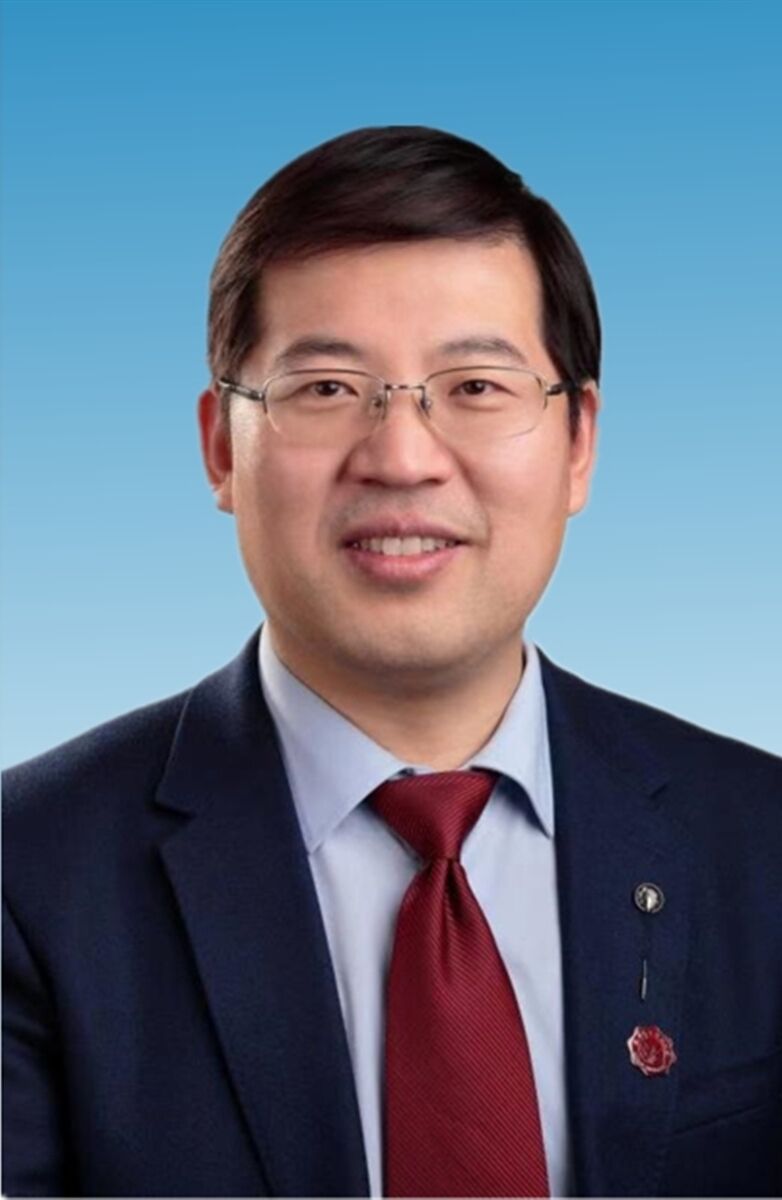}}]{Xinbo Gao}
	(Senior Member, IEEE) received the B.Eng., M.Sc. and Ph.D. degrees in electronic engineering, signal and information processing from Xidian University, Xi'an, China, in 1994, 1997, and 1999, respectively. From 1997 to 1998, he was a research fellow at the Department of Computer Science, Shizuoka University, Shizuoka, Japan. From 2000 to 2001, he was a post-doctoral research fellow at the Department of Information Engineering, the Chinese University of Hong Kong, Hong Kong. Since 2001, he has been at the School of Electronic Engineering, Xidian University. He is a Cheung Kong Professor of Ministry of Education of P. R. China, a Professor of Pattern Recognition and Intelligent System of Xidian University. Since 2020, he is also a Professor of Computer Science and Technology of Chongqing University of Posts and Telecommunications. His current research interests include Image processing, computer vision, multimedia analysis, machine learning and pattern recognition. He has published six books and around 300 technical articles in refereed journals and proceedings. Prof. Gao is on the Editorial Boards of several journals, including Signal Processing (Elsevier) and Neurocomputing (Elsevier). He served as the General Chair/Co-Chair, Program Committee Chair/Co-Chair, or PC Member for around 30 major international conferences. He is a Fellow of the Institute of Engineering and Technology and a Fellow of the Chinese Institute of Electronics.
\end{IEEEbiography}

\end{document}